\documentclass[sigconf]{acmart}
\usepackage{algorithm}
\usepackage{algorithmicx}
\usepackage{algpseudocode}
\usepackage{amsmath}
\usepackage{amsthm}
\usepackage{textcomp}
\usepackage{multirow}
\usepackage{hyperref}
\usepackage{diagbox}
\usepackage{url}
\usepackage{enumitem}
\usepackage{makecell}
\usepackage{tabularx} 
\usepackage{subcaption}
\newtheorem{theorem}{Theorem}

\newtheorem{definition}{Definition}[theorem]
\newtheorem{proposition}{Proposition}

\AtBeginDocument{%
  \providecommand\BibTeX{{%
    \normalfont B\kern-0.5em{\scshape i\kern-0.25em b}\kern-0.8em\TeX}}}
\setcopyright{acmcopyright}
\copyrightyear{2024}
\acmYear{2024}

\acmConference[KDD'24]{KDD'24: SIGKDD Conference on Knowledge Discovery and Data Mining}{August 25-29, 2024}{Barcelona, Spain}
\acmBooktitle{KDD'24: SIGKDD Conference on Knowledge Discovery and Data Mining, August 25-29, 2024, Barcelona, Spain}

\begin{document}

\title{Cross-Context Backdoor Attacks against Graph Prompt Learning}
\author{Xiaoting Lyu}
\affiliation{%
  \institution{School of Computer Science and Technology \\Beijing Jiaotong University}
  \city{Beijing}
  \country{China}
}
\email{xiaoting.lyu@bjtu.edu.cn}

\author{Yufei Han}
\affiliation{%
  \institution{Inria, Univ. Rennes, IRISA}
  \country{France}}
\email{yufei.han@inria.fr}

\author{Wei Wang}
\authornote{Corresponding author}
\affiliation{%
  \institution{School of Computer Science and Technology \\Beijing Jiaotong University}
  \city{Beijing}
  \country{China}
}
\email{wangwei1@bjtu.edu.cn}

\author{Hangwei Qian}
\affiliation{%
 \institution{CFAR, A*STAR}
 \country{Singapore}}
\email{qian_hangwei@cfar.a-star.edu.sg}

\author{Ivor Tsang}
\affiliation{%
 \institution{CFAR, A*STAR}
 \country{Singapore}}
\email{ivor_tsang@cfar.a-star.edu.sg}

\author{Xiangliang Zhang}
\affiliation{%
  \institution{University of Notre Dame}
  \state{Indiana}
  \country{USA}}
\email{xzhang33@nd.edu}
\begin{abstract}

Graph Prompt Learning (GPL) bridges significant disparities between pretraining and downstream applications to alleviate the knowledge transfer bottleneck in real-world graph learning. While GPL offers superior effectiveness in graph knowledge transfer and computational efficiency, the security risks posed by backdoor poisoning effects embedded in pretrained models remain largely unexplored. Our study provides a comprehensive analysis of GPL's vulnerability to backdoor attacks. We introduce \textit{CrossBA}, the first cross-context backdoor attack against GPL, which manipulates only the pretraining phase without requiring knowledge of downstream applications. 
Our investigation reveals both theoretically and empirically that tuning trigger graphs, combined with prompt transformations, can seamlessly transfer the backdoor threat from pretrained encoders to downstream applications.
Through extensive experiments involving 3 representative GPL methods across 5 distinct cross-context scenarios and 5 benchmark datasets of node and graph classification tasks, we demonstrate that \textit{CrossBA} consistently achieves high attack success rates while preserving the functionality of downstream applications over clean input. 
We also explore potential countermeasures against \textit{CrossBA} and conclude that current defenses are insufficient to mitigate \textit{CrossBA}. Our study highlights the persistent backdoor threats to GPL systems, raising trustworthiness concerns in the practices of GPL techniques.

\end{abstract}



\maketitle
\section{Introduction}

Real-world graph learning tasks pose challenges in generalization and knowledge transfer when deploying pretrained graph neural networks (GNNs) to downstream applications divergent from the pretraining stage. For instance, a GNN pretrained on social networks may be utilized in recommendation systems, while encoders designed for link prediction might be repurposed for node or graph classification tasks. 
The substantial differences between pretraining and downstream applications, including variations in problem domains, semantic space, and learning objectives \cite{allinone,graphprompt,sun2023graph}, present obstacles for transferring the graph knowledge in pretrained GNN models to diverse downstream applications. 
In response, Graph Prompt Learning (GPL) \cite{allinone,graphprompt,sun2023graph,VNT} has emerged as a promising solution for such graph learning tasks requiring the generalization of graph knowledge across various application contexts (abbreviated as cross-context graph learning). 
Inspired by prompt learning in Large Language Models (LLMs) \cite{DBLP:journals/csur/LiuYFJHN23,Rogers_Kovaleva_Downey_Rumshisky_2020}, GPL involves training GNN encoders initially on unannotated pretext graph data and then tailoring prompts for downstream applications to guide these encoders. This approach effectively bridges the gap between pretraining and downstream tasks without altering the GNN's parameters, thereby avoiding resource-intensive data annotation and model retraining while facilitating robust generalization of pretrained GNN encoders across diverse downstream applications.

While GPL facilitates knowledge transfer across diverse graph learning tasks in cross-context scenarios, it also exposes downstream applications to the risk of inheriting backdoors embedded in pretrained models. Attackers can implant backdoors into pretrained GNN encoders, leading to downstream models built on these encoders inheriting the backdoor poisoning effects and misclassifying backdoored inputs to attacker-desired target labels. Such backdoor vulnerabilities \cite{NOTABLE, UOR, DBLP:conf/iclr/ChenMSG0LF22,Zhao2023prompt,cai2022badprompt,yao2023poisonprompt} have been identified in prompt learning in Natural Language Processing (NLP), which involves using rare words as triggers and associating them with specific target classes or output embeddings. However, adapting these NLP-based attacks to graph learning encounters challenges due to fundamental differences in data structure and learning paradigms.

Prior research \cite{DBLP:conf/sacmat/ZhangJWG21,DBLP:conf/ccs/XuP22,DBLP:conf/colcom/ShengCCK21,DBLP:conf/ijcai/XuC0CWHL19,DBLP:conf/uss/XiPJ021} has revealed vulnerabilities of GNNs to backdoor attacks, manipulating labeled graph data during supervised training to induce misclassification. However, these attacks are not applicable to GPL scenarios. Unlike traditional GNN tasks, GPL constructs GNN encoders using unlabeled data during unsupervised pretraining, limiting attackers' access to labeled training data for injecting backdoor noise. Recent work \cite{GCBA} targeting graph contrastive learning (GCL) associates triggers with target class embeddings to cause misclassification. However, this method requires prior knowledge about downstream applications, rendering it infeasible in cross-context GPL scenarios where attackers only control the pretraining process without information about downstream applications. Additionally, existing research fails to investigate the generalization and transferability of backdoor attacks across different cross-context GPL scenarios. In summary, organizing successful backdoor attacks against cross-context GPL systems remains largely unexplored.


\noindent
\textit{\textbf{Presented Work.} } We study the feasibility of transferable backdoor attacks against various GPL methods across diverse cross-context scenarios. Building on prior research \cite{allinone,graphprompt,sun2023graph,DBLP:conf/aaai/ZhangPYZC023,DBLP:journals/corr/abs-2209-15240,DBLP:journals/corr/abs-2310-14845,VNT}, we categorize cross-context learning into 5 scenarios: cross-task, cross-domain, cross-dataset, cross-class, and cross-distribution, as detailed in Section \ref{sec:preliminaries}, based on the disparities between pretraining and downstream graph data. These scenarios offer a comprehensive assessment of the inherent backdoor threats to GPL.

Realizing such attacks in the aforementioned cross-context GPL scenarios poses unique challenges compared to conventional graph backdoor attacks.
\emph{First}, attackers in cross-context GPL scenarios can only access and control the unlabeled graph data collected during pretraining. They lack awareness of downstream applications where the model may be deployed, rendering traditional backdoor attacks unfeasible.
\emph{Second}, the divergences between pretraining and downstream applications in semantic spaces, structural patterns, and learning objectives challenge the Independent and Identically Distributed (IID) assumption fundamental to machine learning models, hindering the generalization of the backdoor poisoning effects embedded in the pretrained model.
\emph{Third}, successful backdoor attacks should ensure that the backdoored model remains functional on clean input in downstream applications. Additionally, the attack should retain its effectiveness even when countermeasure mechanisms are deployed by downstream users.

To this end, we propose \textit{CrossBA}, the first cross-context backdoor attack method against GPL, which addresses the aforementioned challenges from the following perspectives. 

\textit{First of all}, \textit{CrossBA} formulates the backdoor attack as a feature collision-oriented optimization problem during the pretraining stage. \textit{CrossBA} is designed to simultaneously associate backdoored graphs with the embedding of the trigger graph and ensure that these embeddings are distinct from those of clean graphs. In this way, any backdoored graph will be mapped to the target embedding, causing downstream applications to misclassify it as the target class determined by the target embedding.
\textit{Furthermore}, our theoretical analysis in Section \ref{sec:Theoretical} unveils the intrinsic vulnerability of GPL systems to backdoor poisoning effects embedded in pretrained GNN models. While prompt learning facilitates knowledge transfer to downstream applications, it inadvertently amplifies the transferability of backdoors. Additionally, both our theoretical analysis and empirical observations demonstrate that optimizing the trigger graph to align the loss landscape of the backdoor and main learning task can further enhance the transferability of backdoors in cross-context GPL scenarios.
\textit{Finally}, to enhance the stealthiness of the attack, \textit{CrossBA} optimizes the trigger graph to align the loss landscape of the backdoor task with that of the main task, while also reinforcing the closeness between the embeddings of the backdoor and backdoor-free GNN encoders on the same clean input graphs, thereby minimizing utility loss in downstream applications.
Moreover, \textit{CrossBA} constrains the node features of the trigger graph to closely resemble those of clean nodes, aiding in evading potential countermeasures deployed by downstream users.

We evaluate \textit{CrossBA} against 3 representative GPL methods, ProG \cite{allinone}, ProG-Meta \cite{allinone} and GraphPrompt \cite{graphprompt}, using 5 real-world graph datasets for both graph and node classification tasks. Our evaluation covers 5 various cross-context graph learning scenarios, including \emph{cross-distribution}, \emph{cross-class}, \emph{cross-dataset}, \emph{cross-domain}, and \emph{cross-task}, as detailed in Section \ref{sec:preliminaries}. 
Despite challenges, \textit{CrossBA} consistently achieves attack success rates exceeding 0.85 across various downstream applications in all 5 cross-context scenarios, while maintaining high utility with less than a 0.06 drop in classification accuracy compared to backdoor-free counterparts.
Comparisons with GCBA \cite{GCBA}, a relevant backdoor attack method for GCL, reveal \textit{CrossBA}'s superior performance across diverse cross-context scenarios. For instance, against the GAT model trained by ProG on CiteSeer, \textit{CrossBA} outperforms {GCBA} by at least 68\% in both accuracy over clean data and attack success rate. 
Additionally, we discuss potential countermeasures against \textit{CrossBA} and conclude that the current defenses in GPL are insufficient to mitigate \textit{CrossBA}, yet effective against GCBA. For example, in cross-domain scenarios, \textit{CrossBA} achieves attack success rates above 0.90 facing PruneG, while those of the baselines drop below 0.50.

Our contributions focus on answering the 3 research questions:

\noindent\textbf{RQ1:} How does an attacker organize successful cross-context backdoor attacks against GPL systems?

\noindent\textbf{RQ2:} Do different GPL methods exhibit equal susceptibility to cross-context backdoor attacks? Is there a shared underlying cause for the vulnerability of various GPL methods to such attacks?

\noindent\textbf{RQ3:} Does the threat of backdoor attacks persist across different cross-context GPL scenarios?


To address \textbf{RQ1}, we begin by providing an overview of relevant literature and foundational concepts of the GPL framework in Sections \ref{sec:related} and \ref{sec:preliminaries}. We then delve into the threat model of cross-context backdoor attacks and the design of \textit{CrossBA} in Sections \ref{sec:threat model} and \ref{sec:crossba}.
For \textbf{RQ2}, we theoretically analyze the feasibility of \textit{CrossBA} within the GPL framework in Section \ref{sec:Theoretical}, revealing the inherent backdoor vulnerability in GPL.  
To bolster the findings for \textbf{RQ1} and \textbf{RQ2}, and to address \textbf{RQ3}, we conduct a comprehensive empirical evaluation in Section \ref{sec:evaluation} to demonstrate the persistent threat posed by \textit{CrossBA} across diverse cross-context scenarios and against different GPL methods. Section \ref{sec:conclusion} concludes the entire paper.




\section{Related works}

\noindent
\textit{\textbf{Graph Prompt Learning.}}
Prompt learning, initially successful in NLP \cite{DBLP:conf/acl/GaoFC20}, has been extended to graph data, tailoring prompts for downstream tasks to guide the pretrained model to perform effectively without altering its parameters. Prompts in graph learning manifest in two forms: prompt as tokens and prompt as graphs \cite{sun2023graph,fang2023universal,DBLP:journals/corr/abs-2310-15318,allinone,graphprompt,DBLP:journals/corr/abs-2310-17394,DBLP:journals/corr/abs-2309-10131}. Two representative methods of these GPL methods are GraphPrompt \cite{graphprompt} and ProG \cite{allinone}. Both unify pretraining and downstream tasks into a common template but differ in prompts. ProG uses learnable, graph-structured variables as prompts attached to the input graph, while GraphPrompt embeds prompt tokens as learnable vectors into the hidden layers of the GNN model, enhancing the Readout operation.

\noindent
\textit{\textbf{Backdoor Attacks.}} 
Backdoor attacks on GNNs have gained attention recently \cite{DBLP:conf/sacmat/ZhangJWG21,DBLP:conf/ccs/XuP22,DBLP:conf/colcom/ShengCCK21,DBLP:conf/ijcai/XuC0CWHL19,DBLP:conf/uss/XiPJ021,Dai2023www}. In these attacks, the backdoored GNN model predicts an attacker-chosen label for any testing input embedded with triggers. 
Notably, \cite{DBLP:conf/sacmat/ZhangJWG21} employs the Erdos-Rényi (ER) model to generate subgraphs as triggers, \cite{DBLP:conf/uss/XiPJ021} introduces an adaptive trigger generator enhancing attack effectiveness, \cite{DBLP:conf/ccs/XuP22} designs triggers that preserve the labels of backdoored samples, and \cite{DBLP:conf/ijcai/XuC0CWHL19} proposes unnoticeable graph backdoor attacks to bypass defense mechanisms.
However, in cross-context GPL scenarios, traditional graph backdoor attacks are inapplicable, as the attacker only controls the pretraining process with unlabeled data. 
GCBA \cite{GCBA}, targeting GCL, manipulates the victim GNN encoder to associate a trigger with the target class's embedding. Yet, GCBA necessitates knowledge of the target class in downstream tasks, impractical in cross-context GPL scenarios.
Our study explores the feasibility of backdoor attacks in cross-context GPL scenarios, where attackers can only use unlabeled data to pretrain GNN encoders. Importantly, attackers cannot access or interfere with downstream applications.

\label{sec:related}

\section{Preliminaries}
We focus on the workflow of cross-context graph prompt learning \cite{allinone,graphprompt}. Attackers pretrain the GNN encoder using self-supervised learning on unlabeled graph data. Downstream users then learn the prompts with few-shot training samples based on the pretrained encoder. Relevant concepts and definitions are introduced below.


\noindent
\textit{\textbf{GNN encoder.}}  GNNs have become a predominant approach for learning graph embeddings. Typically, GNNs utilize a neighborhood aggregation strategy, wherein the encoder iteratively updates a node's embedding by aggregating embeddings from its neighbors through message passing. Formally, at the $k$-th layer, the embedding of node $v_i$ is given by:
\begin{equation}
\small
h_{{v_i}}^k = \text{AGGREGATE}\left(h_{{v_i}}^{k - 1}, \{v_j \in \mathcal{N}({v_i}) \cup {v_i}\}, \theta^{(k)}\right)
\label{eq:aggregate}
\end{equation}
where $\mathcal{N}({v_i})$ is the set of first-order neighbors of node $v_i$ in the graph $G$, and the AGGREGATE function combines neighborhood node embeddings to update the node embedding. 
The final node embedding of $v_i$ is denoted as $h_{v_i}=E_{\theta}(G,v_i)$, where $\theta$ denotes the parameters of the encoder $E$. The graph embedding $E_{\theta}(G)$ is then obtained through a Readout function that aggregates node embeddings from the entire graph.
The objective functions of pretraining include various self-supervised tasks such as GraphCL \cite{graphcl} and link prediction \cite{graphprompt}, which enable the model to capture rich structural and feature-based graph patterns.
Our study focuses on inductive learning, where the GNN encoder's input is the inductive graph of a given node, including the node itself and its k-hop neighbors.

\noindent
\textit{\textbf{Graph prompt learning.}} This study is involved in two sophisticated GPL methods: ProG \cite{allinone} and GraphPrompt \cite{graphprompt}. Both ProG and GraphPrompt unify pretraining and downstream tasks into graph-level tasks, introducing learnable prompts to guide these tasks.
In ProG, the prompt graph is denoted as $G_{pro}=({P,S})$, where $P=\{v^p_1,v^p_2,\dots,v^p_{|P|}\}$ represents the set of $|P|$ nodes, each characterized by a token vector ${{\rm{p}}_i} \in {\mathbb{R}^{1 \times d}}$. The set $S=\{(v^p_i,v^p_j)|v^p_i,v^p_j \in P\}$ defines the prompt graph's topology structure. 
A prompted graph is obtained by inserting the prompt graph $G_{pro}$ into the input graph $G$. The parameters for both the prompt graph and the answering function of downstream tasks are optimized through few-shot learning. ProG-Meta enhances ProG by incorporating meta-learning techniques.
GraphPrompt introduces prompts within the hidden layers of the GNN model to assist the graph pooling operation. Given the node set $V=\{v_1,v_2,\dots,v_N\}$ in $G$ with each node's embedding $h_{v_i}$, and a learnable prompt vector $\rm{p}_j$ for the downstream task $j$, the prompt-assisted readout operation for graph $G$ is defined as a reweighed readout function, $ {\rm{Readout(\{ }}{{\rm{p_j}}} \odot {h_{{v_i}}}|{v_i} \in {V\} )}$, where $\odot$ represents element-wise multiplication. The prompt vector $\rm{p}_{j}$ updates its parameters by gradient descent to minimize the graph similarity loss.

\noindent
\textit{\textbf{Cross-context few-shot learning.}} Cross-context few-shot learning is designed to facilitate rapid adaptation of models to new tasks in diverse application contexts using limited labeled examples.
The inherent data heterogeneity in this paradigm results in disparities in both semantic space and data distribution between pretraining and downstream contexts. We summarize the cross-context scenarios studied in previous works \cite{allinone,graphprompt,VNT,sun2023graph,DBLP:journals/corr/abs-2209-15240,DBLP:journals/corr/abs-2310-14845} and delineate five cross-context scenarios to cover different levels of such disparities:
\begin{itemize}[leftmargin=*]
\item \textit{Cross-task:} This setting reflects the divergence in the goal of graph learning tasks. The pretraining   may be conducted for  classification of an entire graph, such as ENZYMES, while the downstream task involves classification of nodes, like CiteSeer.
\item \textit{Cross-domain:} The pretraining and downstream data originate from distinct domains but share the same task type. For instance, pretraining could be conducted on commercial product networks (e.g., Amazon), while downstream tasks involve academic citation networks (e.g., Cora), both focusing on node classification tasks.
\item \textit{Cross-dataset:} This setting involves different datasets within the same domain. For instance, the pretraining is on dataset like Cora, while the downstream task involves a different dataset like CiteSeer. Both are academic citation networks but  different.
\item \textit{Cross-class:} Both the pretraining and downstream datasets stem from the same data source, such as Cora. However, they focus on different classes, i.e.,    pretraining and downstream tasks have different class distributions. .
\item \textit{Cross-distribution:} 
In this scenario, both the pretraining and downstream datasets are sourced from the same data origin, sharing identical features and label spaces. However, they present distinctly different data distributions.
\end{itemize}

\label{sec:preliminaries}
\section{Threat Models}
\label{sec:threat model}

\begin{figure*}[t] 
\centering  
\includegraphics[height=5cm,width=17.5cm]{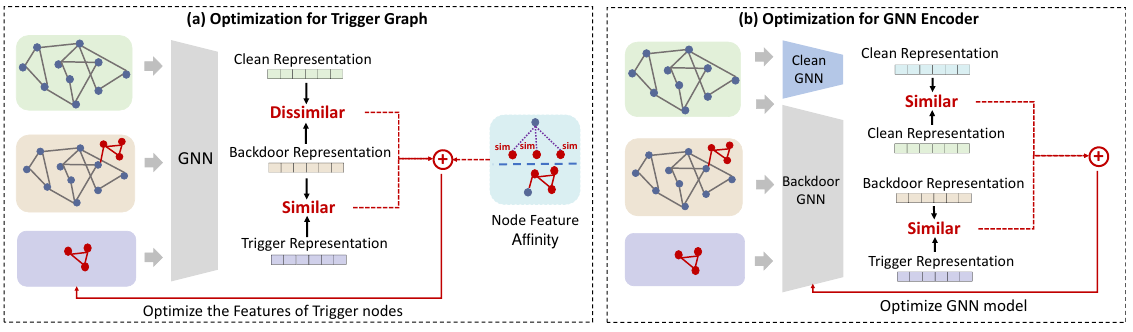} \caption{The  attack workflow of CrossBA.} 
\label{fig:attackflow}
\end{figure*}

 \noindent
\textit{\textbf{Attacker's goal.}} Attackers aim to mislead downstream models built upon the attacker-crafted GNN encoder to classify backdoored inputs as the target class. Simultaneously, the downstream model should behave normally for clean inputs. Specifically, in cross-context scenarios, attackers build the backdoored GNN encoder at the pretraining phase, which memorizes the association between backdoored inputs and the attacker-desired output embedding. This backdoored GNN encoder is then adapted for downstream applications using the GPL methods. 
To ensure the transferability of backdoors in cross-context GPL, the attack should be organized with the following properties: 1) \textit{Context-agnostic}: The backdoor attack should remain consistently effective across various downstream contexts. 2) \textit{Prompt-agnostic}: The attack should be adaptable to different designs of graph prompt learning methods used by downstream users. 3) \textit{Stealthy}: The backdoored GNN model should contribute close classification accuracy to backdoor-free models on clean data of downstream applications. Furthermore, the attack remains effective with the defense mechanisms deployed by downstream users.




 \noindent
\textit{\textbf{Attacker's capability.}}
We assume that the attacker possesses complete control over the pretraining process. The attacker can inject backdoor triggers into the unlabeled pretraining data and access the training methods for the backdoored GNN encoder. However, the attacker is incapable of accessing or manipulating the labeled data and the GPL training process utilized by downstream users.

 \noindent
\textit{\textbf{Attacker's knowledge.}}
The attacker has full knowledge of the pretraining phase, including the dataset, the architecture of the pretrained GNN encoder, and the training method. On the contrary,  the attacker lacks any information regarding the GPL training process conducted by downstream users, including specifics about the downstream datasets utilized and the GPL methods applied.


\section{Cross-context Backdoor Attack}
\label{sec:crossba}

\subsection{Overview of CrossBA}
Figure \ref{fig:attackflow} provides an overview of \textit{CrossBA}. 
In our attack scenario, attackers train the backdoored GNN encoder using unlabeled graph data and self-supervised methods like graph contrastive learning. 
\textit{CrossBA} aims to inject embedding collisions between backdoored graphs and the trigger graph into the backdoored GNN encoder, while maintaining a distinct embedding for the backdoored graph compared to its clean counterpart. 
This backdoor poisoning effect causes the backdoored graph to be associated with a target embedding distant from the clean input, leading to misclassification in downstream models. 
Additionally, \textit{CrossBA} also optimizes the standard contrastive learning objective over clean graphs to ensure robust classification performance.



Formally, the trigger graph serves as the backdoor signal is $\Delta_G=(V_{\Delta}, E_{\Delta})$, with $V_{\Delta}=\{v_\Delta^1,\dots,v_\Delta^C\}$ representing the set of $C$ trigger nodes. Each trigger node has a feature vector ${\rm{x}}_\Delta^i \in {\mathbb{R} ^{1 \times d}}$, matching the input graph's node feature dimension. $E_{\Delta} = \{ ({v_{\Delta}^i},{v_{\Delta}^j})|{v_{\Delta}^i},{v_{\Delta}^j} \in V_{\Delta}\} $ defines the links of the trigger graph. To integrate the trigger graph $\Delta_G$ into the input graph $G$ for generating the backdoored graph, the attacker randomly selects a node in $G$ as the anchor node $v_{att}$,  linking it to a specific node $v_\Delta^i$ in the trigger graph. We limit the trigger graph to a 3-node and fully connected graph, significantly smaller than graphs in both pretraining and downstream datasets.

\subsection{Attack Objective of CrossBA}\label{sec:attackobjective}
We define the objective function of the \textit{CrossBA} attack   in Eq.\ref{eq:optimise}, which jointly optimizes the backdoored GNN encoder parameters ${\theta_b}$ and the node features of the trigger graph $\Delta_G$. 
\begin{equation}
\small
\begin{array}{c}
\theta _b^*,\Delta _G^* = \mathop {\arg \min }\limits_{{{{\theta _b}}},{\Delta _G}} {{\mathcal L}_{{\rm{bdk}}}}{\rm{ + }}{{\mathcal L}_{{\rm{clr}}}} + \alpha {\mathcal L}_{{\rm{sim}}}^c + \beta {\mathcal L}_{{\rm{sim}}}^a,\\
{\rm{where}}\;\;\;{\kern 1pt} {{\mathcal L}_{{\rm{bdk}}}} =  - \frac{1}{{|D|}}\sum\limits_{{G_i} \in D} {\rm{sim} } \left( {{{\hat E}_{{\theta _b}}}({G_i} \oplus {\Delta _G}),{{\hat E}_{{\theta _b}}}({\Delta _G})} \right)\\
 + \lambda \frac{1}{{|D|}}\sum\limits_{{G_i} \in D} {\rm{sim} } \left( {{{\hat E}_{{\theta _b}}}({G_i} \oplus {\Delta _G}),{{\hat E}_{{\theta _b}}}({G_i})} \right),\\
{{\mathcal L}_{{\rm{clr}}}} =  - \frac{1}{{\left| D \right|}}\sum\limits_{{G_i} \in D} {\log \frac{{\exp ({\rm{sim}}({H_{{G_i}}},{H_{G_i^ + }})/\tau )}}{{\exp ({\rm{sim}}({H_{{G_i}}},{H_{G_i^ + }})/\tau ) + \sum\limits_{k \ne i} {\exp ({\rm{sim}}({H_{{G_i}}},{H_{G_{{k}}^ - }})/\tau )} }}} ,\\
{\mathcal L}_{{\rm{sim}}}^c =  - \frac{1}{{|D|}}\sum\limits_{{G_i} \in D} {\rm{sim} } \left( {{{\hat E}_{{\theta _b}}}({G_i}),{E_{{\theta _c}}}({G_i})} \right),\\
{\mathcal L}_{{\rm{sim}}}^a =  - \frac{1}{{|D|}}\sum\limits_{{G_i} \in D} {\sum\limits_{{v_\Delta^j} \in {V_\Delta}} {\rm{sim} } } \left( {{\rm{feat}}({v_\Delta^j}),{\rm{feat}}({v_{{{att}}}})} \right).
\end{array}
\label{eq:optimise}
\end{equation}
where $\hat{E}_{\theta_b}$ represents the backdoored GNN encoder, with $\theta_b$ denoting its parameters. $G_i$ is a clean graph in the pretraining dataset $D$. $G_i^+$ is an augmented graph of $G_i$ obtained by randomly adding or removing links, while $G_k^-$ is a graph  different from $G_i$ in $D$. The operation $\oplus$ attaches the trigger graph $\Delta_G$ into the anchor node $v_{att}$ of the clean graph $G$. $\mathrm{feat}(v_i)$ denotes the feature vector of a node $v_i$. $E_{\theta_c}$ is the backdoor-free GNN encoder trained with clean graph data. 
$H_{G_i}=\hat E_{{\theta _b}}(G_i)$ denotes the embedding of the clean graph $G_i$. ${\rm{sim}} (u,v)$ is the similarity between the embeddings $u$ and $v$. $\tau $ is the temperature parameter. The parameters $\lambda$, $\alpha$, and $\beta $ balance the weight of the loss terms.

\vspace{+0.05in}
\noindent
\textit{\textbf{The main learning loss $\mathcal{L}_{\rm clr}$}} defines a contrastive learning objective of the main task as in \cite{graphcl}. The goal is to train a GNN encoder capable of producing distinctive embeddings for clean graph data. Specifically, we maximize the similarity between the embeddings of the graph $G_i$ and its augmented counterpart ${G}_i^+$. Simultaneously, we maximize the dissimilarity between the embedding of any other graph ${G}_k^-$ and that of $G_i$.


\vspace{+0.05in}
\noindent
\textit{\textbf{The backdoor learning loss ${\mathcal L_{{\rm{bdk}}}}$ }} defines the backdoor task's objective. To address the attacker's lack of knowledge about downstream datasets, \textit{CrossBA} utilizes the embedding of the trigger graph as the target, and induces the backdoor mapping into the GNN encoder by colliding the embeddings of backdoored graphs with that of the trigger graph. By jointly tuning the backdoored GNN encoder and the trigger graph, we ensure that the embeddings of backdoored graphs resemble the target embedding while being different from clean graphs' embeddings. Consequently, when applied to downstream applications, backdoored graphs will be misclassified to the target class associated with the target embedding.

Compared to manually specifying a static backdoor mapping (with a fixed trigger graph and target embedding), our design offers several advantages. 
\emph{First}, by jointly tuning the target embedding and trigger graph, we align the loss landscapes of the main task and the backdoor task within the fixed GNN encoder. This minimizes the backdoor learning loss while preserving the utility of the backdoored GNN encoder on clean graph data.
\emph{Second}, fine-tuning the trigger graph, as supported by our theoretical analysis in Section \ref{sec:Theoretical}, reduces the disparity in backdoor task performance between pretraining and downstream applications, thereby enhancing backdoor transferability. 
\emph{Lastly}, we parameterize the tuning of the target embedding by optimizing the trigger graph, ensuring it remains within the embedding space of the graphs in $D$. Directly optimizing the target embedding as an independent variable may lead to extreme values outside the span of graphs in $D$, making them prone to detection by downstream anomaly detection methods.

\vspace{+0.05in}
\noindent
\textit{\textbf{The embedding alignment loss $\mathcal{L}_{\mathrm{sim}}^c$ }} is designed to ensure that backdoors do not impair the GNN encoder's ability to generate discriminative embeddings for clean graph data. The attacker can build a clean GNN encoder $E_{\theta_c}$ using clean pretraining data as a reference model. By aligning the output embeddings of $E_{\theta_c}$ with those of $\hat{E}_{\theta_b}$ on the same clean inputs, the backdoored GNN encoder $\hat{E}_{\theta_b}$ can perform similarly to the clean encoder $E_{\theta_c}$ on clean data.
  
\vspace{+0.05in}
\noindent
\textit{\textbf{The node feature affinity loss $\mathcal L_{{\rm{sim}}}^{{a}}$}} is designed to enhance the stealthiness of \textit{CrossBA} to evade anomaly detection-based sanitary checks over node features. Since adjacent nodes in a graph typically share similar features, downstream users can check the node feature consistency to identify abnormal nodes with significantly deviated feature values from their neighbors \cite{GCBA,dai2023unnoticeable}. To circumvent such defenses, we optimize the node features of the trigger graph by minimizing $\mathcal L_{{\rm{sim}}}^{{a}}$, enforcing feature consistency between the nodes in the trigger graph and the anchor nodes in the clean input graph. 

\subsection{Alternating Optimization for CrossBA}
Algorithm \ref{alg:attackxxx} in Appendix \ref{app:alg} outlines the procedural flow of the \textit{CrossBA} attack.
Initially, attackers employ self-supervised methods, such as GraphCL \cite{graphcl}, to train a clean GNN encoder ${E}_{\theta_c}$ by minimizing $\mathcal{L}_{\rm clr}$. 
The trigger injection is then conducted by optimizing the attack objective function in Eq. \ref{eq:optimise} in alternating order. During each attack round, attackers first freeze the GNN encoder $\hat{E}_{\theta_b}$ and optimize the node features of the trigger graph $\Delta_G$. Subsequently,  attackers optimize the backdoored GNN encoder $\hat{E}_{\theta_b}$ based on the optimized trigger graph $\Delta_G^*$ and the clean GNN encoder ${E}_{\theta_c}$.

\vspace{+0.05in}
\noindent
\textbf{\textit{Tuning trigger graph.}}
After injecting the trigger graph $\Delta_G$ into the input graph $G$, the attacker optimizes the node features of $\Delta_G$, potentially optimizing the target embedding, by minimizing the backdoor learning loss ${\mathcal L_{{\rm{bdk}}}}$ and the node feature affinity loss $\mathcal L_{{\rm{sim}}}^{{a}}$ with respect to the fixed GNN encoder $\hat{E}_{\theta_b}$.  For simplicity, the update of the trigger node features in one step is given by: 
\begin{equation}
\small
\Delta _G^t = \Delta _G^{t - 1} - \gamma_t {\nabla _{\Delta _G^{t - 1}}}({\mathcal L_{{\rm{bdk}}}}+\beta \mathcal L_{{\rm{sim}}}^{{a}})
\label{eq:trigger_update}
\end{equation}

\vspace{+0.05in}
\noindent
\textbf{\textit{Tuning backdoored GNN encoder.} }
Upon completing trigger optimization, the attacker connects the optimized trigger graph $\Delta_G$ to the anchor node in $G$, recreating backdoored graphs. The optimization aims to maximize the similarity between the embeddings of backdoored graphs and the trigger graph, as well as the similarity of the clean graph's embeddings between $\hat E_{\theta_b}$ and $E_{\theta_c}$. For simplicity, the GNN encoder parameters are updated accordingly:
\begin{equation}
\small
\begin{array}{c}
\theta _b^t = \theta _b^{t - 1} - {\gamma _g}{\nabla _{\theta _b^{t - 1}}}( - \frac{1}{{|D|}}\sum\limits_{{G_i} \in D} {\rm{sim} } \left( {{{\hat E}_{{\theta _b}}}({G_i} \oplus {\Delta _G^t}),{{\hat E}_{{\theta _b}}}({\Delta _G})} \right)\\
 - \alpha \frac{1}{{|D|}}\sum\limits_{{G_i} \in D} {\rm{sim} } \left( {{{\hat E}_{{\theta _b}}}({G_i}),{E_{{\theta _c}}}({G_i})} \right))
\end{array}
\label{eq:gnn_update}
\end{equation}

\subsection{Attack Feasibility of CrossBA}\label{sec:Theoretical}
In this section, we explore the feasibility of the proposed \textit{CrossBA} attack against cross-context GPL. To simplify the analysis, we adopt the prompt graph setting from \cite{allinone}, where the prompt graph $G_{pro}$ is attached to the input graph $G$. We define $D_{s}$ and $D_t$ as the distributions for the pretraining and downstream graph datasets, respectively. Downstream users create the prompted graph by $G_{t} \otimes {G_{{pro}}}$.  The encoder outputs the embedding of the prompted graph as $\hat{h}(G_{t} \otimes {G_{{pro}}}) = \hat{E}_{\theta_b}(G_{t} \otimes {G_{{pro}}})$.

\begin{theorem}\label{theorem:feasibility} 
Assuming a $L_{\hat{E}}$-Lipschitz continuous GNN encoder $\hat{E}_{\theta_b}$ with $k_{G}$-node input graphs, and the node feature matrix $X$ of a graph $G=(V,E)$ has a bounded Frobenius norm, i.e., $|X|_{fro}\leq{\mu}$.
Upon freezing the GNN encoder $\hat{E}_{\theta_b}$, the backdoor learning loss $\mathcal{L}_{\rm bdk}$ in Eq.\ref{eq:optimise} is upper-bounded by the main learning loss $\mathcal{L}_{\rm clr}$ at the pretraining stage, as given in Eq. \ref{eq:upperbound_pretrain}:
\begin{equation}
\small
\begin{array}{cc}
\mathcal{L}^{s}_{\rm bdk} \leq \mathcal{L}^{s}_{\rm clr} + 2\sqrt{d(\Delta^{+}_G,G_{s}\oplus \Delta_G)} + C\\
d(\Delta^{+}_G,G_{s}\oplus \Delta_G) = 2\sqrt{k_{G}}(k_{G}-1)L_{\hat{E}}\mu - \sum\limits_{i= 1}^n \sum\limits_{j = 1}^m \frac{s(\hat{h}(\Delta^{+}_{G,i}),\hat{h}(G_{s,j} \oplus \Delta_{G}))}{nm}
\end{array}
\label{eq:upperbound_pretrain}
\end{equation}
Similarly, the main learning loss in the downstream context can be upper bounded by the main learning loss in the pretraining context:
\begin{equation}\label{eq:upperbound_main}
\small
\begin{array}{cc}
\mathcal{L}^{t}_{\rm clr} \leq \mathcal{L}^{s}_{\rm clr} + 2\sqrt{d(G_{s}, G_{t} \otimes {G_{{pro}}})} + C_{0}\\
d(G_{s}, G_{t} \otimes {G_{{pro}}}) = {2\sqrt{k_{G}}(k_{G}-1)L_{\hat{E}}\mu - \sum\limits_{i= 1}^n \sum\limits_{j = 1}^m \frac{s(\hat{h}(G_{s,i}),\hat{h}(G_{t,j}  \otimes  {G}_{{pro}}))}{nm}} 
\end{array}
\end{equation}
Combing Eq.\ref{eq:upperbound_pretrain} and Eq.\ref{eq:upperbound_main}, the backdoor learning loss in the downstream context can be upper bounded as in Eq.\ref{eq:upperbound_bdk}: 
{\begin{equation}\label{eq:upperbound_bdk}
\small
\begin{array}{*{20}{l}}
{{\mathcal L}_{{\rm{bdk}}}^t \le {\mathcal L}_{{\rm{clr}}}^s}
{ + {{(8\sqrt {{k_G}} ({k_G} - 1){L_{\hat E}}\mu  - 4\sum\limits_{i = 1}^n {\sum\limits_{j = 1}^m {\frac{{s(\hat h(\Delta _{G,i}^ + ),\hat h({G_{s,j}} \oplus {\Delta _G}))}}{{nm}}} } )}^{1/2}}}\\
{ + {{(8\sqrt {{k_G}} ({k_G} - 1){L_{\hat E}}\mu  - 4\sum\limits_{i = 1}^n {\sum\limits_{j = 1}^m {\frac{{s(\hat h({G_{s,i}} \oplus {\Delta _G}),\hat h(({G_{t,j}} \oplus {\Delta _G}) \otimes {G_{{{pro}}}}))}}{{nm}}} } )}^{1/2}}} + {C_1}
\end{array} 
\end{equation}}\normalsize
where $\Delta_{G}^+$ are augmented trigger graphs with randomly added or removed links from $\Delta_G$. 
$G_{s}$ and $G_{t}$ are the graphs sampled from the pretraining and downstream data distribution $D_{s}$ and $D_{t}$, respectively. The RKHS kernel function $s(*,*)$ measures the similarity between two graph embeddings.
$C$, $C_0$, and $C_1$ are constants for the GNN encoder's complexity and the optimal learning loss of an ideal encoder.

\end{theorem}

\noindent\textbf{Observation 1. Transferability of graph prompt learning v.s. transferability of backdoor attacks.}
\begin{proposition}\label{proposition1}
Given a backdoored GNN encoder $\hat{E}_{\theta_b}$, there exists a prompt graph ${G}_{{pro}}$ that maximizes the similarity values $\sum\limits_{i= 1}^n \sum\limits_{j = 1}^m \frac{s(\hat{h}(G_{s,i}),\hat{h}(G_{t,j}  \otimes  {G}_{{pro}}))}{nm}$ between the graphs from the pretraining dataset and the prompted graphs from the downstream dataset. 
\end{proposition}
By combining Proposition \ref{proposition1}, Eq.\ref{eq:upperbound_main}, and Eq.\ref{eq:upperbound_bdk}, we discover that employing prompts in the cross-context GPL presents both advantages and disadvantages. On one hand, augmenting downstream graph data from $D_t$ with the prompt $G_{pro}$ enhances the adaptability of pretrained GNN encoders to downstream tasks by alleviating distributional disparities between $D_t$ and $D_s$ within the GNN encoder's embedding space. This results in a well-trained GNN encoder capable of generating discriminative embeddings for new tasks. However, on the other hand, as indicated by Eq.\ref{eq:upperbound_bdk}, incorporating the prompt graph into the backdoored graphs from $D_{t}$ inadvertently facilitates backdoor attack transferability by reducing the upper bound of the backdoor learning loss in downstream tasks.



\vspace{+0.03in}
\noindent\textbf{Observation 2. Tuning the trigger graph enhances the transferability of backdoor poisoning effects while concurrently preserving the utility of backdoored GNN models.}
\begin{proposition}\label{proposition2}
Given a backdoored GNN encoder $\hat{E}_{\theta_b}$, there exists a trigger graph $\Delta_{G}$ that maximizes the similarity measure $\frac{1}{{nm}}{\sum\limits_{i= 1}^n \sum\limits_{j = 1}^m s(\hat{h}(\Delta^{+}_{G,i}),\hat{h}(G_{s,j} \oplus \Delta_{G}))}$ between the augmented variants of the trigger graph and the backdoored graphs in  pretraining.
\end{proposition}
In \textit{CrossBA}, we propose optimizing the trigger graph $\Delta_G$ with a fixed GNN encoder during the pretraining stage to minimize the backdoor learning loss.  This process, as indicated by Eq.\ref{eq:upperbound_pretrain} and Proposition \ref{proposition2}, improves the alignment between the embeddings of $\Delta^{+}_G$ and $G_{s}\oplus{\Delta_G}$, narrowing the disparity between the main learning loss and the backdoor learning loss on pretraining data. 
This ensures that given a well-trained GNN encoder, the trigger tuning module reduces the backdoor learning loss without compromising the performance of the main task. 
Furthermore, as shown in Eq.\ref{eq:upperbound_bdk}, tuning the trigger graph during pretraining, along with the prompt graph, further facilitates backdoor attack transferability by lowering the upper bound of the backdoor learning loss in downstream tasks, leading to misclassification with backdoored input.

The theoretical investigation elucidates the feasibility of delivering cross-context backdoor attacks following the design of \textit{CrossBA}, providing a response to \textbf{RQ1}.  Moreover, Observation 1, which addresses \textbf{RQ2}, reveals the dual nature of knowledge transfer within GPL's prompt learning. While prompt learning enhances downstream models with the pretrained model's expertise, it also poses the risk of backdoor transfer. Our findings underscore substantial concerns regarding the trustworthiness of GPL methods. Proofs of Theorem \ref{theorem:feasibility} and Propositions \ref{proposition1} and \ref{proposition2} are provided in Appendix \ref{app:proof}.

\section{Experimental Evaluation}
\label{sec:evaluation}

\begin{table*}[htbp]
\centering
\footnotesize
\caption{ACC, ASR, and AD in cross-distribution scenarios.  }
\label{tab:cross_distribution}
\resizebox{\linewidth}{!}{
\begin{tabular}{@{}ccc *{12}{|c}@{}}
\toprule
& & & \multicolumn{8}{c|}{\textbf{Node Classification}} & \multicolumn{4}{c}{\textbf{Graph Classification}} \\ 
\cmidrule(lr){4-11} \cmidrule(lr){12-15}
\multirow{2}{*}{\textbf{GPL}} & \multirow{2}{*}{\textbf{Model}} & \multirow{2}{*}{\textbf{Attack}} & \multicolumn{2}{c|}{\textbf{CiteSeer}} & \multicolumn{2}{c|}{\textbf{Cora}} & \multicolumn{2}{c|}{\textbf{Computers}} & \multicolumn{2}{c|}{\textbf{Photo}} &\multicolumn{2}{c|}{\textbf{CiteSeer-Graph}}&\multicolumn{2}{c}{\textbf{Photo-Graph}} \\ 
\cmidrule(lr){4-5} \cmidrule(lr){6-7} \cmidrule(lr){8-9} \cmidrule(lr){10-11} \cmidrule(lr){12-13}  \cmidrule(lr){14-15}  
 & & & ACC(AD) & ASR & ACC(AD) & ASR& ACC(AD) & ASR& ACC(AD) & ASR& ACC(AD) & ASR & ACC(AD) & ASR\\ 
\midrule
\multirow{6}{*}{ProG} & \multirow{3}{*}{GAT} & GCBA\_R   & 0.24 (+0.60) & 0.51 &  0.33(+0.32) &0.76 & 0.43(+0.29) & 0.18 & 0.61(+0.18) &0.15 & 0.17(+0.54) & 0.00& 0.66(+0.23) & 0.24\\
                          &                  & GCBA\_M   &  0.24 (+0.60) & 0.62 & 0.18(+0.47) & 0.00 & 0.40(+0.32) & 0.44 &  0.65(+0.14) & 0.18&  0.17(+0.54) & 0.00 & 0.69(+0.20) & 0.15\\
                          && CrossBA &\textbf{ 0.83(+0.01)} & \textbf{0.90} & \textbf{0.64} (+0.01) & \textbf{1.00} &  \textbf{0.70(+0.02)} &\textbf{ 1.00} &  \textbf{0.79(-0.00)}  & \textbf{0.94}& \textbf{0.76(-0.05)}&\textbf{ 1.00} & \textbf{0.89(-0.00)}& \textbf{0.91}\\
\cmidrule(lr){2-15}
                          & \multirow{3}{*}{GT} & GCBA\_R   & 0.35(+0.47) & 0.96 &  0.35(+0.38) & 0.69 & 0.47(+0.27) & 1.00 & 0.57(+0.22) &  0.85 & 0.37(+0.42) & 0.32 & 0.58(+0.33) &  0.89\\
                          &                     & GCBA\_M   & 0.35(+0.47) &0.96 & 0.45(+0.28) & 0.23 &0.41(+0.33) &0.75 & 0.60(+0.19) & 0.30 &0.33(+0.46)  & 0.21& 0.57(+0.34) & 0.56\\
                          && CrossBA & \textbf{0.82(-0.00)} & \textbf{1.00} & \textbf{0.73(-0.00)} & \textbf{0.99} & \textbf{0.72(+0.02)} & \textbf{1.00} & \textbf{0.79(-0.00)} &\textbf{1.00} & \textbf{0.80(-0.01)} & \textbf{1.00} &\textbf{0.92(-0.01)} &  \textbf{0.99}\\
\midrule
\multirow{6}{*}{\begin{tabular}[c]{@{}l@{}}Graph\\Prompt\end{tabular}}   & \multirow{3}{*}{GAT} & GCBA\_R   & 0.69(+0.07) &0.05 &  0.47(+0.11)&0.14 & 0.49(+0.11) & 0.14 & 0.59(+0.07) &  0.00 & 0.72(-0.00) & 0.04& 0.58(+0.20) & 0.00\\
                          &                          & GCBA\_M   & 0.78(-0.02) & 0.01 & 0.43(+0.15) &  0.07  & 0.43(+0.17) &0.13 & 0.56(+0.10)& 0.15 & 0.67(+0.05) & 0.13& 0.51(+0.27) & 0.28\\
                          && CrossBA & \textbf{0.80(-0.04) }& \textbf{1.00} & \textbf{0.67(-0.09)} & \textbf{1.00} & \textbf{0.62(-0.02)}& \textbf{1.00}&  \textbf{0.68(-0.02)} & \textbf{0.99} & \textbf{0.72(-0.00)} &\textbf{0.99}& \textbf{0.74(+0.04)} & \textbf{1.00}\\
\cmidrule(lr){2-15}
                          & \multirow{3}{*}{GT} & GCBA\_R   &  0.69(+0.09) & 0.16 & 0.56(+0.15) & 0.19  & 0.50(+0.15) & 0.15 & 0.62(+0.13) &  0.13 & 0.66(+0.09) & 0.10 &0.55(+0.29) &0.25\\
                          &                     & GCBA\_M   & 0.64(+0.14) & 0.04 & 0.53(+0.18) & 0.25 & 0.47(+0.18) & 0.10 & 0.59(+0.16)& 0.12 & 0.74(+0.01) & 0.11& 0.57(+0.27) & 0.14\\
                          && CrossBA &  \textbf{0.79(-0.01)} & \textbf{1.00} & \textbf{0.70(+0.01)} & \textbf{0.99} & \textbf{0.67(-0.02)} & \textbf{1.00} & \textbf{0.74(+0.01)} & \textbf{1.00} & \textbf{0.77(-0.02)} & \textbf{1.00}& \textbf{0.84(-0.00)} & \textbf{1.00}\\
\midrule[0.75pt] 
\multirow{6}{*}{\begin{tabular}[c]{@{}l@{}}ProG\\Meta\end{tabular}}  & \multirow{3}{*}{GAT}  & GCBA\_R   &  \textbf{0.85(-0.17)} &  0.56 & 0.50(+0.38) & 0.00 & 0.57(+0.25) & 0.00 &  0.75(+0.24) & 0.88 & 0.50(+0.38)& 0.00 & 0.86(+0.14)  & 0.94\\
                          &                        & GCBA\_M   &  0.50(+0.18) & 0.00 & 0.50(+0.38) & 0.00  &  0.57(+0.25) & 0.00 & 0.50(+0.49) &  0.00 &  0.50(+0.38) & 0.00& 0.98(+0.02) & 0.00\\
                          && CrossBA & 0.84(-0.16) & \textbf{1.00} & \textbf{0.89(-0.01)} & \textbf{1.00} & \textbf{0.79(+0.03)} & \textbf{1.00} & \textbf{0.99(-0.00)} & \textbf{1.00} & \textbf{0.85(+0.03)} & \textbf{1.00}& \textbf{1.00(-0.00)} & \textbf{1.00}\\
\cmidrule(lr){2-15}
                          & \multirow{3}{*}{GT} & GCBA\_R   & 0.93(-0.02) &  0.90 & 0.90(+0.01) &  0.31 &0.57(+0.29)  & 0.00 &0.97(+0.02) & 0.71 & 0.92(+0.03) & 0.05 & 0.50(+0.50) & 0.00\\
                          &                     & GCBA\_M   &  0.50(+0.41) & 0.00 & 0.88(+0.03) &  0.05 & 0.57(+0.29) &  0.00 & 0.99(-0.00) &  0.10 & 0.87(+0.08) & 1.00 & 1.00(-0.00) & 0.98\\
                          && CrossBA & \textbf{0.93(-0.02)} & \textbf{1.00} &  \textbf{0.90(+0.01)} &  \textbf{1.00} & \textbf{0.85(+0.01)} & \textbf{1.00} & \textbf{0.99(-0.00)} & \textbf{ 1.00}& \textbf{0.94(+0.01)} &\textbf{1.00}& \textbf{1.00(-0.00)} & \textbf{1.00}\\
\bottomrule
\end{tabular}
}
\end{table*}


\subsection{Experimental Settings}
\noindent
\textit{\textbf{Datasets and the Backbone GNN Models.}} We utilize 5 benchmark datasets for evaluation: CiteSeer \cite{yang2016revisiting}, Cora \cite{yang2016revisiting}, Amazon-Computers \cite{shchur2018pitfalls}, Amazon-Photo \cite{shchur2018pitfalls}, and ENZYMES \cite{borgwardt2005protein}. CiteSeer, Cora, Amazon-Computers, and Amazon-Photo are utilized for node classification tasks, while ENZYMES is employed for graph classification tasks.  Furthermore, following the setting in \cite{allinone}, we define graph classification tasks using the node classification datasets. In GPL systems, we employ two advanced GNN models: Graph Attention Network (GAT) \cite{GAT} and Graph Transformer (GT) \cite{GT}. Details about datasets and GNN models can be found in Appendix \ref{app:datasets}.

\noindent
\textit{\textbf{GPL Methods.}} We evaluate the attack performance against mainstream GPL methods applicable to both node and graph classification tasks, categorized into two types \cite{sun2023graph}: \emph{Prompt as Graphs} and \emph{Prompt as Tokens}. For the former branch, we choose ProG \cite{allinone} and ProG-Meta \cite{allinone}, formulating prompts as subgraph patterns.
For the latter, we target GraphPrompt \cite{graphprompt}, which considers prompts as tokens added to the Readout module of GNNs. 

\noindent
\textit{\textbf{Baseline Attacks.}} 
GCBA \cite{GCBA} emerges as the most relevant backdoor attack for our study, considering the threat model. While not explicitly tailored for GPL, GCBA aims to inject backdoor poisoning noise into a GNN encoder trained via GCL. 
In GCBA, the attacker gathers graph data of the target class in downstream applications and utilizes the GCBA-crafting method to inject the backdoor into the GNN encoder.
However, GCBA does not directly apply to cross-context GPL scenarios as it necessitates access to downstream applications.
We introduce two variants of GCBA adapted to our threat model: {GCBA\_R} and {GCBA\_M}.  In both variants, the attacker initially clusters the embeddings of clean data collected during the pretraining stage, utilizing the backdoor-free GNN encoder. Subsequently, GCBA\_R randomly selects the embedding at the center of a cluster as the target embedding, while GCBA\_M chooses the cluster center embedding farthest from other clusters as the target embedding. Further details can be found in Appendix \ref{app:baselineattack}.

\noindent
\textit{\textbf{Evaluation Metrics.} }
We employ 3 metrics to evaluate attack effectiveness: (1) Attack Success Rate (ASR) \cite{GCBA}, representing  the accuracy with which a backdoored downstream model classifies backdoored inputs to the target class designated by the target embedding, (2) Accuracy of the Main Task (ACC), measuring the classification accuracy of backdoored downstream models over clean graph data, and (3) Accuracy Drop (AD) \cite{GCBA}, denoting the difference in ACC between backdoor-free and backdoored downstream models. A lower AD indicates less utility loss of backdoored GNN encoders. A successful backdoor attack should ensure high ACC (low AD) and high ASR values simultaneously.


\noindent

\noindent\textit{\textbf{Implementations.}} For the details about the implementations, including the cross-context scenarios, the GPL methods, and all the attacks, please refer to Appendix \ref{app:implementation}. We provide codes in the link \footnote{https://github.com/xtLyu/CrossBA}.

\begin{table*}[htbp]
\centering
\footnotesize
\caption{ACC, ASR, and AD in cross-class scenarios.  }
\label{tab:cross_class}
\resizebox{\linewidth}{!}{
\begin{tabular}{@{}ccc *{12}{|c}@{}}
\toprule
& & & \multicolumn{8}{c|}{\textbf{Node Classification}} & \multicolumn{4}{c}{\textbf{Graph Classification}} \\ 
\cmidrule(lr){4-11} \cmidrule(lr){12-15}
\multirow{2}{*}{\textbf{GPL}} & \multirow{2}{*}{\textbf{Model}} & \multirow{2}{*}{\textbf{Attack}} & \multicolumn{2}{c|}{\textbf{CiteSeer}} & \multicolumn{2}{c|}{\textbf{Cora}} & \multicolumn{2}{c|}{\textbf{Computers}} & \multicolumn{2}{c|}{\textbf{Photo}} &\multicolumn{2}{c|}{\textbf{CiteSeer-Graph}}&\multicolumn{2}{c}{\textbf{Photo-Graph}} \\ 
\cmidrule(lr){4-5} \cmidrule(lr){6-7} \cmidrule(lr){8-9} \cmidrule(lr){10-11} \cmidrule(lr){12-13}  \cmidrule(lr){14-15}  
 & & & ACC(AD) & ASR & ACC(AD) & ASR & ACC(AD) & ASR& ACC(AD) & ASR & ACC(AD) & ASR & ACC(AD) & ASR\\ 
\midrule
\multirow{6}{*}{ProG} & \multirow{3}{*}{GAT} & GCBA\_R   & 0.25(+0.68) & 0.00 & 0.47(+0.31) & 1.00 & 0.55(+0.28) & 1.00 & 0.73(+0.06) & 0.21 & 0.26(+0.62) & 0.00 & 0.80(+0.12) & 0.84\\
                          &                           & GCBA\_M   & 0.25(+0.68) & 0.00 & 0.49(+0.29) & 1.00 & 0.56(+0.27) & 0.95& 0.68(+0.11) &  0.36 & 0.47(+0.41) & 1.00& 0.79(+0.13) & 1.00 \\
                          && CrossBA & \textbf{0.93(-0.00)} & \textbf{1.00} & \textbf{0.78(-0.00)} & \textbf{1.00} & \textbf{0.83(-0.00)} &\textbf{1.00} & \textbf{0.79(-0.00)} & \textbf{1.00} & \textbf{0.87(+0.01)} & \textbf{1.00}&  \textbf{0.93(-0.01)} & \textbf{1.00} \\
\cmidrule(lr){2-15}
                          & \multirow{3}{*}{GT} & GCBA\_R   & 0.49(+0.44)& 0.29 &0.61(+0.27) & 0.99 & 0.60(+0.22) & 0.72 & 0.62(+0.13) & 1.00 & 0.49(+0.39) & 0.46 & 0.76(+0.10) & 0.01\\
                          &                           & GCBA\_M   & 0.49(+0.44) & 0.34 &  0.52(+0.36) &  1.00 & 0.70(+0.12) & 1.00 &  0.69(+0.06) & 0.71 & 0.66(+0.22)& 0.26 &  0.80(+0.06) & 0.80\\
                          && CrossBA &  \textbf{0.93(-0.00)} & \textbf{1.00} & \textbf{0.88(-0.00)} &\textbf{ 1.00} & \textbf{0.82(-0.00)}& \textbf{1.00} & \textbf{0.76(-0.01)} &\textbf{ 1.00} &  \textbf{0.88(-0.00)} & \textbf{1.00} & \textbf{0.88(-0.02) }&  \textbf{1.00}\\
\midrule
\multirow{6}{*}{\begin{tabular}[c]{@{}l@{}}Graph\\Prompt\end{tabular}}  & \multirow{3}{*}{GAT}  & GCBA\_R   & 0.92(-0.04) &  0.05 & 0.86(-0.01) &  0.07 & 0.69(+0.06) &0.10 &  \textbf{0.74(-0.04)}& 0.01 & \textbf{0.87(-0.03) }&0.06 & 0.61(+0.17) & 0.28\\
                          &                           & GCBA\_M   & 0.89(-0.01) & 0.02 &  0.81(+0.04)&0.03 & 0.70(+0.05) & 0.06 & 0.60(+0.10) & 0.00 & 0.84(-0.00) &0.06 & 0.67(+0.11) & 0.27\\
                          && CrossBA & \textbf{0.92(-0.04)} & \textbf{1.00} &\textbf{0.87(-0.02)} &\textbf{ 1.00} & \textbf{0.77(-0.02)} & \textbf{1.00 }&{ 0.72(-0.02)} & \textbf{0.94}& 0.83(+0.01) & \textbf{1.00} &\textbf{ 0.79(-0.01)} & \textbf{1.00}\\
\cmidrule(lr){2-15}
                          & \multirow{3}{*}{GT} &GCBA\_R   & 0.87(+0.06) & 0.03&  0.76(+0.11)& 0.03 & 0.72(+0.08) &  0.08 & 0.66(+0.12) & 0.03 & 0.88(-0.01) & 0.05 & 0.59(+0.21) & 0.24\\
                          &                           & GCBA\_M   & 0.88(+0.05) & 0.05 & 0.83(+0.04) & 0.00 & 0.70(+0.10) &  0.05& 0.66(+0.12) & 0.15 & 0.87(-0.00) & 0.04 &  0.66(+0.14) & 0.15\\
                          && CrossBA & \textbf{0.93(-0.00)} &\textbf{ 1.00} & \textbf{0.88(-0.01)} & \textbf{1.00} &\textbf{0.80(-0.00)} & \textbf{1.00} &  \textbf{0.79(-0.01)} & \textbf{1.00} & \textbf{0.89(-0.02)} & \textbf{1.00} & \textbf{0.87(-0.07)} &\textbf{1.00}\\
\midrule[0.75pt] 
\multirow{6}{*}{\begin{tabular}[c]{@{}l@{}}ProG\\Meta\end{tabular}}  & \multirow{3}{*}{GAT} & GCBA\_R   &  0.50(+0.30) & 0.00 & 0.50(+0.14) & 0.00 & 0.57(+0.24) & 0.00 &  0.99(-0.00) &  1.00 & 0.50(+0.42) & 0.00 & 1.00(-0.00) & 1.00\\
                          &                           & GCBA\_M   & 0.50(+0.30) & 0.00 & 0.50(+0.14) & 0.00 & 0.57(+0.24) & 0.00 & 0.76(+0.23) & 1.00 & 0.50(+0.42) & 0.00 &  0.88(+0.12) & 1.00\\
                          && CrossBA & \textbf{0.75(+0.05)} & \textbf{1.00} & \textbf{0.80(-0.16)} & \textbf{1.00} & \textbf{0.80(+0.01)} & \textbf{1.00}&\textbf{0.99(-0.00)} & \textbf{ 1.00} &  \textbf{0.94(-0.02)} &  \textbf{1.00} & \textbf{1.00(-0.00)} & \textbf{1.00}\\
\cmidrule(lr){2-15}
                          & \multirow{3}{*}{GT} & GCBA\_R   & 0.50(+0.45) &  0.00 & 0.52(+0.35) & 0.00 & 0.57(-0.28) & 0.00 &  0.99(+0.01)&  1.00 &0.50(+0.44) & 0.00 & 0.87(+0.13) & 0.38\\
                          &                           & GCBA\_M   & 0.77(+0.18) & 1.00 &  0.60(+0.27) &  1.00 & 0.57(-0.28) & 0.00 &  0.99(+0.01) & 1.00 & 0.58(+0.36) &0.02 & 0.94(+0.06)& 1.00\\
                          && CrossBA & \textbf{0.95(-0.00)} & \textbf{1.00} &  \textbf{0.84(+0.03)}& \textbf{1.00} & \textbf{0.85(-0.00)} & \textbf{1.00} &\textbf{ 1.00(-0.00)}&\textbf{ 1.00} & \textbf{0.95(-0.01)} &\textbf{ 1.00 }&\textbf{ 1.00(-0.00)} & \textbf{1.00}\\
\bottomrule
\end{tabular}
}
\end{table*}

\begin{table*}[htbp]
\centering
\footnotesize
\caption{ACC, ASR, and AD in cross-domain scenarios.  Photo is used as the pretraining dataset.}
\label{tab:cross_application}
\begin{tabular}{@{}ccc *{8}{|c}@{}}
\toprule
& & & \multicolumn{4}{c|}{\textbf{Node Classification}} & \multicolumn{4}{c}{\textbf{Graph Classification}} \\ 
\cmidrule(lr){4-7} \cmidrule(lr){8-11}
\multirow{2}{*}{\textbf{GPL}} & \multirow{2}{*}{\textbf{Model}} & \multirow{2}{*}{\textbf{Attack}} & \multicolumn{2}{c|}{\textbf{CiteSeer}} & \multicolumn{2}{c|}{\textbf{Cora}} & \multicolumn{2}{c|}{\textbf{CiteSeer-Graph}} & \multicolumn{2}{c}{\textbf{Cora-Graph}} \\ 
\cmidrule(lr){4-5} \cmidrule(lr){6-7} \cmidrule(lr){8-9} \cmidrule(lr){10-11}
 & & & ACC(AD) & ASR & ACC(AD) & ASR & ACC(AD) & ASR & ACC(AD) & ASR \\ 
\midrule
\multirow{6}{*}{ProG} & \multirow{3}{*}{GAT}          & GCBA\_R   & 0.24(+0.50) & 1.00 & 0.18(+0.40) & 1.00 & 0.31(+0.41) & 0.80 & 0.14(+0.74) & 0.00 \\
                          &                           & GCBA\_M   & 0.20(+0.54) & 0.00 & 0.32(+0.26) & 1.00 & 0.20(+0.52) & 0.88 & 0.14(+0.74) &  0.00 \\
                          && CrossBA     & \textbf{0.76(-0.02)} &\textbf{1.00} &\textbf{ 0.73(-0.15)} &\textbf{1.00} & \textbf{0.72(-0.00)} & \textbf{1.00} &  \textbf{0.85(+0.03)} & \textbf{1.00} \\
\cmidrule(lr){2-11}
                          & \multirow{3}{*}{GT}       & GCBA\_R   &0.40(+0.43) & 1.00 & 0.34(+0.39) & 0.00 & 0.33(+0.47) & 1.00 & 0.14(+0.76) & 0.00 \\
                          &                           & GCBA\_M   &  0.42(+0.41) & 1.00 & 0.33(+0.40) & 1.00 & 0.44(+0.36) & 1.00 & 0.40(+0.50)& 1.00 \\
                          && CrossBA     & \textbf{0.83(-0.00)}& \textbf{1.00} &\textbf{ 0.73(-0.00)} & \textbf{1.00} & \textbf{0.80(-0.00)} & \textbf{1.00 }& \textbf{0.90(-0.00)} &\textbf{ 1.00} \\
\midrule
\multirow{6}{*}{\begin{tabular}[c]{@{}l@{}}Graph\\Prompt\end{tabular}}  & \multirow{3}{*}{GAT} & GCBA\_R   &  0.75(-0.03) & 0.85 & 0.68(-0.00) & 0.24 & 0.59(+0.15) & 0.00 & 0.78(+0.05) & 0.60 \\
                          &                           & GCBA\_M   &  \textbf{0.77(-0.05)} & 0.71 & 0.58(+0.10) &0.00& 0.63(+0.11) & 0.03 & 0.64(+0.19) & 0.23 \\
                          && CrossBA & 0.76(-0.04) & \textbf{1.00} &\textbf{ 0.70(-0.02)} &  \textbf{1.00} &\textbf{ 0.74(-0.00)} &\textbf{1.00} &\textbf{ 0.81(+0.02)} &\textbf{1.00 }\\
\cmidrule(lr){2-11}
                          & \multirow{3}{*}{GT}  & GCBA\_R   & 0.64(+0.15) & 0.19 & 0.57(+0.14) &  0.02 & 0.68(+0.08) & 0.00 & 0.74(+0.10) & 0.04 \\
                          &                           & GCBA\_M   & 0.70(+0.09) & 0.00 & 0.53(+0.18) & 0.00 & 0.58(+0.18) &0.00 & 0.76(+0.08)& 0.00 \\
                          & & CrossBA     & \textbf{0.79(-0.00)} &\textbf{0.99} & \textbf{0.72(-0.01) }& \textbf{0.99} & \textbf{0.77(-0.01)} &\textbf{ 1.00 }& \textbf{0.82(+0.02)} & \textbf{1.00} \\
\midrule[0.75pt] 
\multirow{6}{*}{\begin{tabular}[c]{@{}l@{}}ProG\\Meta\end{tabular}}  & \multirow{3}{*}{GAT} & GCBA\_R   & 0.50(+0.34) & 0.89 & 0.50(+0.39) & 0.03 & 0.50(+0.38) &0.00 & 0.50(+0.26) & 0.00\\
                          &                           & GCBA\_M   & 0.50(+0.34) & 0.00 & 0.50(+0.39) &0.00 & 0.50(+0.38) &1.00 & 0.50(+0.26) & 0.00\\
                          & & CrossBA & \textbf{0.80(+0.04) }& \textbf{1.00} & \textbf{0.88(+0.01)} &\textbf{1.00} &  \textbf{0.90(-0.02)} & \textbf{1.00} & \textbf{0.87(-0.11)} & \textbf{1.00} \\
\cmidrule(lr){2-11}
                          & \multirow{3}{*}{GT}      & GCBA\_R   &  0.91(+0.02) & 1.00 & 0.83(+0.07) & 0.52 & 0.84(+0.08) & 1.00& \textbf{0.96(+0.01)} & 1.00 \\
                          &                           & GCBA\_M   & 0.50(+0.43) & 0.00 & 0.50(+0.40) & 1.00 & 0.50(+0.42) & 0.00 &  0.50(+0.47) & 0.00 \\
                          & & CrossBA     & \textbf{0.93(-0.00)} & \textbf{1.00} &  \textbf{0.90(-0.00)} & \textbf{1.00} & \textbf{0.94(-0.02)} & \textbf{1.00} & 0.95(+0.02) &\textbf{1.00} \\
\bottomrule
\end{tabular}
\end{table*}

\begin{figure*}[t] 
\centering  
\includegraphics[height=3cm,width=16.5cm]{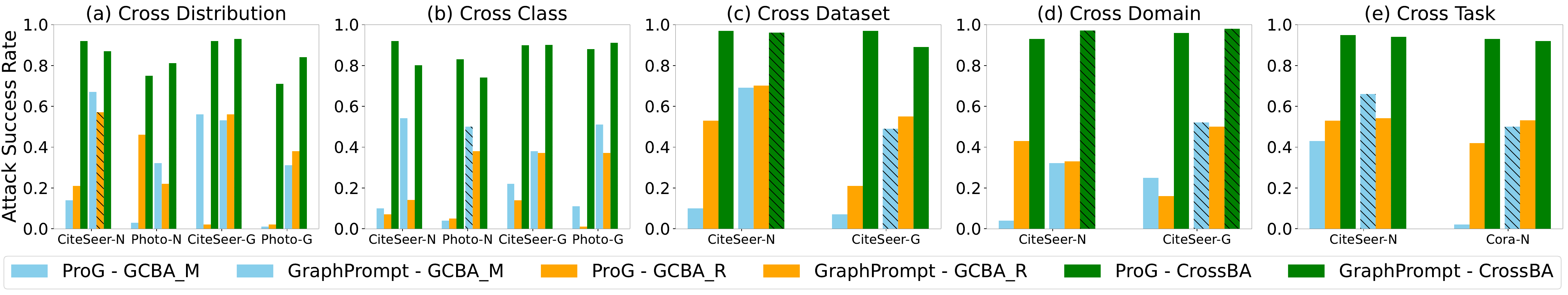} \caption{ASR of backdoor attacks against PruneG in 5 cross-context scenarios. "-N" denotes the node classification task, while "-G" represents the graph classification task.} 
\label{fig:defense}
\end{figure*}

\subsection{Experimental Results}
\textit{\textbf{(1) Attack Performance in Cross-Context GPL Scenarios.}}
We evaluate the feasibility of backdoor attacks on node and graph classification tasks across 5 cross-context scenarios using 3 mainstream GPL methods, addressing 3 research questions. Tables \ref{tab:cross_distribution}--\ref{tab:cross_application} and \ref{tab:cross_dataset}--\ref{tab:cross_level} present the ACC and ASR of all attack methods in cross-distribution, cross-class, cross-domain, cross-dataset, and cross-task scenarios, respectively. Due to space limitations and consistent trends across different scenarios, we report results for cross-distribution, cross-class, and cross-domain scenarios here and defer those for cross-dataset and cross-task scenarios to Appendix \ref{app:attackresult}. The highest ASR and ACC values, and the lowest AD values achieved among all the attacks, including our \textit{CrossBA}, are highlighted in bold.


\noindent\textbf{Superior attack performance of CrossBA across different cross-context scenarios (RQ1 and RQ3)}. 
The results reveal that \textit{CrossBA} consistently outperforms the baseline methods across 5 cross-context GPL scenarios, achieving the highest ASR values while maintaining comparable ACC levels to those of  backdoor-free models. Specifically, \textit{CrossBA} achieves ASR values exceeding 0.87 in all scenarios, with only a negligible difference in ACC compared to the backdoor-free models, at most 0.06 lower.  In contrast, the baseline attacks fail to achieve comparable ASR to \textit{CrossBA} across various scenarios while maintaining ACC simultaneously. For example, the 2 baseline methods never achieve ASR above 0.32 against the GT model trained by GraphPrompt in all 5 scenarios, and their ASR values against GAT models trained by all 3 GPL methods on CiteSeer are below 0.48 in cross-dataset scenarios. Additionally, when the baselines achieve comparable ASR to \textit{CrossBA}, they suffer a significant drop in ACC. For instance, in cross-class scenarios against ProG, the baseline attacks achieve an ASR of about 0.99 on Cora, similar to \textit{CrossBA}, but with an ACC at least 0.27 lower. Similarly, in cross-distribution scenarios against the GT model trained by ProG, the baseline attacks achieve an ASR of about 0.96, close to \textit{CrossBA}, but with ACC values almost half of \textit{CrossBA}'s.

\textbf{Summary.} \textit{CrossBA} effectively generalizes the backdoor poisoning  effects to diverse cross-context scenarios, outperforming two baseline methods, while maintaining the utility of GNN encoders for downstream tasks. This observation confirms the theoretical principles outlined in Section \ref{sec:Theoretical}, highlighting the efficacy of \textit{CrossBA} in enhancing backdoor transferability in various cross-context scenarios while preserving model utility in downstream tasks.


\noindent\textbf{Consistent attack performance of CrossBA against different GPL methods (RQ2).} \textit{CrossBA} demonstrates robust attack effectiveness across various GPL methods, achieving high ASR while preserving the usability of backdoored models in downstream tasks. Specifically, against all 3 GPL methods, \textit{CrossBA} consistently achieves ASR values above 0.90 in cross-class, cross-distribution, cross-dataset, and cross-domain scenarios. In cross-task scenarios, \textit{CrossBA} achieves ASRs exceeding 0.87. In contrast, baseline attacks fail to deliver consistent attack performance against different GPL methods. For instance, in cross-class scenarios, both baseline methods show ASR values below 0.05 against GAT models trained by all three GPL methods on CiteSeer, indicating a complete failure of the attack. The baseline methods achieve ASR values over 0.99 against ProG on Cora but only reach ASRs of 0.07 for GraphPrompt.

\textbf{Summary.} The results reveal a common vulnerability of \textit{CrossBA} to diverse GPL techniques. This aligns with the theoretical foundations presented in Section \ref{sec:Theoretical}, highlighting the intrinsic risks within the GPL framework that the prompt learning module of GPL facilitates the transfer of backdoors to downstream applications.

\noindent
\textit{\textbf{(2) Potential Countermeasures.}}
We evaluate the resilience of \textit{CrossBA} against potential countermeasures of downstream users. Given the absence of specific defenses tailored for cross-context GPL, we adapt \textit{PruneG} \cite{wu2019adversarial}, originally designed to mitigate adversarial attacks on GNNs. Other defense methods, such as \textit{RandSample} \cite{DBLP:conf/sacmat/ZhangJWG21} and \textit{GNNGuard} \cite{10.5555/3495724.3496501}, are not suitable for cross-context GPL scenarios. This is because both \textit{RandSample} and \textit{GNNGuard} entail training GNN encoders, which is impractical in cross-context GPL. Therefore, we exclusively employ \textit{PruneG} in our evaluation.
\textit{PruneG} is a preprocessing technique that eliminates edges between nodes with dissimilar features, removing components with fewer connected nodes. Figure \ref{fig:defense} illustrates the effectiveness of different attack methods against \textit{PruneG} across 5 cross-context scenarios. \textit{CrossBA} consistently outperforms the baselines, achieving high ASR values exceeding 0.70 against \textit{PruneG}. In cross-domain scenarios, \textit{CrossBA} achieves ASRs above 0.90, while the ASRs of the baselines remain below 0.50. 
Figure \ref{fig:node_sim} in Appendix \ref{app:defense} demonstrates that \textit{CrossBA} ensures trigger node features closely resemble those of clean nodes, achieving indistinguishable similarity for trigger edges, affirming its stealthiness. More details are provided in Appendix \ref{app:defense}.


\noindent
\textit{\textbf{(3) Ablation Study.}}
We conduct ablation studies on \textit{CrossBA} to evaluate the significance of its components. Three variants of \textit{CrossBA} are considered: (1) \textit{CrossBA} without trigger optimization, using a fixed trigger graph and target embedding; (2) \textit{CrossBA} without embedding alignment; (3) \textit{CrossBA} without node feature affinity. The attack performance of \textit{CrossBA} and its variants against 2 GPL methods with \textit{PruneG} across 5 cross-context scenarios is presented in Figure \ref{fig:ablation} in Appendix \ref{app:ablation}. The results show that removing any component significantly reduces ASRs. The ASR value of \textit{CrossBA} remains the highest one compared to the variants. These findings affirm the necessity of integrating trigger graph optimization, embedding alignment-based regularization, and node feature affinity constraint together to achieve successful cross-context backdoor attacks against GPL.

\noindent
\textit{\textbf{(4) Impact of Prompt Tokens.}} Figure \ref{fig:prompt node} in Appendix \ref{app:prompttoken} illustrates how the number of prompt tokens affects the attack performance of \textit{CrossBA} against ProG across 5 cross-context scenarios. The results unveil that the effectiveness of \textit{CrossBA} remains robust, exhibiting little impact from variations in the number of prompt tokens. 

\noindent
\textit{\textbf{(5) Impact of Trigger Nodes.}}
Figure \ref{fig:trigger node} in Appendix \ref{app:triggernodes} illustrates the impact of the number of trigger nodes on \textit{CrossBA}'s attack performance against ProG and GraphPrompt across five different scenarios. The results demonstrate that \textit{CrossBA} consistently achieves ASR values exceeding 0.80, regardless of the number of trigger nodes, highlighting the attack's robustness and efficiency.


\section{Conclusion and Future Work}

In this study, we conduct theoretical and empirical investigations to assess the feasibility of backdoor attacks in cross-context graph prompt learning. Our findings reveal that optimizing trigger graphs, coupled with prompt transformations, significantly enhances backdoor transferability. We introduce \textit{CrossBA}, the first cross-context backdoor attack tailored for GPL, and evaluate its performance across five cross-context scenarios encompassing node and graph classification tasks. Additionally, we explore potential defenses against such attacks. Our results demonstrate that \textit{CrossBA} seamlessly embeds backdoors into various downstream models across diverse cross-context GPL scenarios without compromising main task performance, even under defense deployment. 
However, we acknowledge that our study does not address backdoor attack transferability in textual attribute graphs. As a result, we aim to investigate the transferability of backdoors within textual attribute graphs in the future.

\label{sec:conclusion}

\section{Acknowledgment}
This research is supported in part by Systematic Major Project of China State Railway Group Corporation Limited (No.P2023W002), the French National Research Agency with the reference ANR-23-IAS4-0001 (CKRISP), and the National Research Foundation, Singapore and Infocomm Media Development Authority under its Trust Tech Funding Initiative. Any opinions, findings and conclusions or recommendations expressed in this material are those of the author(s) and do not reflect the views of National Research Foundation, Singapore and Infocomm Media Development Authority.

\bibliographystyle{ACM-Reference-Format}
\bibliography{ref.bib}

\appendix

\section{Optimization Algorithm}\label{app:alg}
The pseudocode for the proposed \textit{CrossBA} attack method is shown in Algorithm \ref{alg:attackxxx}.
\begin{algorithm}[htb]
\small
\caption{CrossBA}
\label{alg:attackxxx}
\begin{algorithmic}[1] 
   \Require
      The pretraining dataset $D=\{ G_1,G_2, \cdots, G_N\}$, the regularization weights $\alpha$, $\beta$, and $\lambda$, the learning rate of the trigger graph $\gamma_t$, the learning rate of the GNN encoder $\gamma_g$, the number of attack rounds $T$. 
  \Ensure The optimized trigger graph ${\Delta_G}$ and the backdoor poisoned GNN encoder model $\hat E_{\theta_b}$.
\State Initialize ${\Delta_G^0}$ and $\theta_c^0$.
\State Train the clean GNN encoder $E_{\theta_c}$ based on $D$ by minimizing $\mathcal{L}_{\rm{clr}}$.
\State $\theta_b^0 \leftarrow \theta_c$, $t \leftarrow 1$.
\While{$t \le T$}
    \State For each graph $G$ in $D$, connect the trigger graph ${\Delta_G^{t-1}}$ to the anchor node to form a backdoored graph $G'$.
    \State Obtain ${\Delta_G^t}$ using Eq.\ref{eq:trigger_update}.
    \State Update the trigger graph embedded in the backdoored graphs with the optimized trigger graph ${\Delta_G^t}$.
    \State Obtain $\theta _b^t$ using Eq.\ref{eq:gnn_update}.
\EndWhile
\end{algorithmic}
\end{algorithm}
In this study, we introduce an optimization framework for joint tuning of the trigger graph and a backdoored GNN encoder. Let $n$ denote the number of training samples, $d$ represent the feature dimensionality, $R$ be the number of training rounds for the clean encoder, and $A$ reflect the time complexity of optimizing the trigger. Initially, the clean GNN encoder is trained for $R$ rounds, with each round having a time complexity of $O(nd^2) $, where the $d^2$ term arises from pairwise feature interactions. Given the $R$ rounds, the initial training phase has a complexity of $O(Rnd^2) $. During the attack phase, the trigger is optimized using all samples with complexity $O(An)$. Subsequently, the optimizer is employed to fine-tune the backdoored GNN encoder. The overall time complexity for the attack process thus becomes $O(Rnd^2+ T(An + nd^2))$, where $T$ represents the number of attack rounds.

\section{Proofs of Theorem and Proposition} \label{app:proof}
\begin{definition}
\textbf{Anchored classifier.} We define a classifier $f$ anchored at an arbitrarily given graph $G_{\text{anchor}}\in D$, where $D$ denotes the distribution of the graph data. $E$ is the graph encoding module of $f$. Given an input graph $G$, the classifier $f_{\text{anchor}}$ is formulated based on the RBF kernel as follows:
\begin{equation}\label{eq:anchored classifier}
\small
f_{\text{anchor}}(G) = exp(-\gamma(\|E(G) - E(G_{\text{anchor}})\|^2)
\end{equation}
where $\gamma$ controls the smoothness of the kernel. $f_{\text{anchor}}$ produces a soft decision output normalized between 0 and 1. A higher/lower output from $f_{\text{anchor}}$ indicates that $G$ has a more/less similar embedding to that of $G_{\text{anchor}}$.
\end{definition}
{ With slight reformulations, the loss functions of the main learning task $\mathcal L_{\text{clr}}$ and the backdoor learning task $\mathcal L_{\text{bdk}}$ in Eq.\ref{eq:optimise} can be unified as a cross-entropy loss of the anchored classifier. This can be expressed as:}
\begin{equation}\label{eq:anchor_loss}
\small
\begin{split}
&\mathcal L = \underset{G\sim{D}}{\mathbb{E}}\,\,\,\underset{G_{\text{anchor}}\sim{D}}{\mathbb{E}}-I(G\in{\{G^{\text{pos}}}_{\text{anchor}}\})f_{G_\text{anchor}}(G) \\
&- I(G\in{\{G^{\text{neg}}_{\text{anchor}}\}})(1-f_{G_\text{anchor}}(G))\\
\end{split}
\end{equation}
where $\{G^{\text{pos}}_{\text{anchor}}\}$ and $\{G^{\text{neg}}_{\text{anchor}}\}$ are the sets of graphs forming positive and negative pairs with the given anchor graph $G_\text{anchor}$.

{$\mathcal L$ simplifies to $\mathcal L_{\text{clr}}$ in Eq.\ref{eq:optimise} if 
\textbf{1)} $\{G^{\text{pos}}_{\text{anchor}}\}$ denotes the set of augmented graphs derived from the trigger graph $G_{\text{anchor}} = \Delta_{G}$ by adding or removing links randomly from $\Delta_G$; 
and \textbf{2)} $G^{\text{neg}}_{\text{anchor}}$ represents any graph sampled from distribution  $D$ that is not in the set of augmented graphs.}

{Similarly, $\mathcal L$ simplifies to $\mathcal L_{\text{bdk}}$ in Eq.\ref{eq:optimise} when  \textbf{1)} $G^{\text{pos}}_{\text{anchor}}$ is defined as $G \oplus \Delta_G$ with $G\in{D}$;   
and \textbf{2)} $G^{\text{neg}}_{\text{anchor}}$ follows the same setting as in $\mathcal L_\text{clr}$, being any graph from $D$ other than $\Delta_G$. }

In this sense, the main learning task aims to minimize $\mathcal L$, with the distribution of input graphs as $D_{G^{+}_{\text{anchor}}}\cup{D}$, where $D_{G^{+}_{\text{anchor}}}$ denotes the distribution of the augmented graphs of $G_{\text{anchor}}$. 
In contrast, the backdoor learning task minimizes $\mathcal L$ over a different distribution of input graphs as $D_{G\oplus{G_{\text{anchor}}}}\cup{D}$. 

\noindent
\textbf{Proof to Theorem 1.} For any anchored classifier $f$, we treat $D_{G^{+}_{\text{anchor}}}\cup{D}$ and $D_{G\oplus{G_{\text{anchor}}}}\cup{D}$ as the data distributions of the source and target problem domain. 
Additionally, within the anchored classifier, we designate the graph encoder $E$ as the backdoored encoder $\hat{E}_{\theta_b}$, which is trained during the pretraining stage.
Following Theorem 1 in \cite{DBLP:conf/icml/LongC0J15}, the backdoor learning loss during the pretraining phase is upper bounded, as shown in Eq.\ref{eq:dd_bdk}:
\begin{equation}\label{eq:dd_bdk}
\small
\begin{split}
&\mathcal L^{s}_{\text{bdk}} \leq \mathcal L^{s}_{\text{clr}} + 2\sqrt{d(\Delta^{+}_{G},{G_{s} \oplus \Delta_G})} + C\\
&d(\Delta^{+}_{G},{G_{s} \oplus \Delta_G}) =\sum\limits_{i,k=1}^n \frac{s(\hat{h}(\Delta^{+}_{G,i}),\hat{h}(\Delta^{+}_{G,k}))}{n^2} \\
&+ \sum\limits_{j,l=1}^m \frac{s(\hat{h}(G_{s,j} \oplus \Delta_{G}),\hat{h}(G_{s,l} \oplus \Delta_{G}))}{m^2} 
- \sum\limits_{i=1}^n \sum\limits_{j=1}^m \frac{s(\hat{h}(\Delta^{+}_{G,i}),\hat{h}(G_{s,j} \oplus \Delta_{G}))}{nm}\\
\end{split}
\end{equation}
where $C$ represents a constant that encapsulates the model complexity of $\hat{E}_{\theta_b}$ and the minimum achievable loss with the ideal parameters for $\hat{E}_{\theta_b}$. $s(,)$ is an RKHS kernel function measuring the similarity between graph embeddings produced by $\hat{E}_{\theta b}$.

{Furthermore, for the simplicity of analysis, we assume the encoder $\hat{E}_{\theta_b}$ is a GIN model defined by: $\hat{h}(G) = (A+(1+\epsilon)I)X{\theta_b}$. 
$A$ denotes the adjacency matrix corresponding to the input graph $G$. The matrix $X$ is a feature matrix, where each row corresponds to the features of node $v_i$ of $G$. ${\theta_b}$ is a multi-layer encoder with a Lipschitz constant denoted by $L_{\hat{E}}$. 
The first two terms are upper bounded independently of the kernel choice. }
\begin{equation}\label{eq:norm}
\small
\begin{split}
&s(\hat{h}(\Delta^{+}_{G,i}),\hat{h}(\Delta^{+}_{G,k})) \leq \|\hat{h}(\Delta^{+}_{G,i})\|\\
&\leq \sqrt{k_{G}}\|\hat{h}(\Delta^{+}_{G,i})\|_{fro} \\
&\leq \sqrt{k_{G}}(k_{G}-1)L_{\hat{E}}\|(A+(1+\epsilon)I)X\|\,\,\,\, \triangleright \text{Lipschitz-continuity condition} \\
&\leq \sqrt{k_{G}}(k_{G}-1)L_{\hat{E}}\|A+(1+\epsilon)I\|_{2}\|X\|_{fro}\\
&\leq \sqrt{k_{G}}(k_{G}-1)L_{\hat{E}}\|X\|_{fro}\\
&\leq \sqrt{k_{G}}(k_{G}-1)L_{\hat{E}}\mu\,\,\,\,\triangleright \text{Bounded frobenius norm}
\end{split}
\end{equation}
where $\|\|$ denotes an abitrary norm of $\hat{h}(\Delta^{+}_{G,i})$. The same bound applies to  $s(\hat{h}(G_{s,j} \oplus \Delta_{G}),\hat{h}(G_{s,l} \oplus \Delta_{G}))$. Integrating Eq.\ref{eq:norm} into Eq.\ref{eq:dd_bdk}, we can derive Eq.\ref{eq:upperbound_pretrain} in Theorem \ref{theorem:feasibility}. 

Similarly, we establish the proof for Eq.\ref{eq:upperbound_main} as follows.
Following Theorem 1 in \cite{DBLP:conf/icml/LongC0J15}, we derive the upper bound for the main learning task in downstream applications, as indicated in Eq.\ref{eq:dd_main}:
{\begin{equation}\label{eq:dd_main}
\small
\begin{split}
&\mathcal L^{t}_{\text{clr}} \leq \mathcal L^{s}_{\text{clr}} + 2\sqrt{d(G_{s},G_{t} \otimes G_{{pro}})} + C_{0}\\
&d(G_{s},G_{t}\otimes G_{{pro}}) = \sum\limits_{i,j=1}^{n}\frac{s(\hat{h}(G_{s,i}), \hat{h}(G_{s,j}))}{n^2} \\
&+ \sum\limits_{i,j=1}^{m}\frac{s(\hat{h}(G_{t,i} \otimes G_{{pro}}), \hat{h}(G_{t,j}\otimes G_{{pro}} ))}{m^2}\\
&- \sum\limits_{i= 1}^n \sum\limits_{j = 1}^m \frac{s(\hat{h}(G_{s,i}),\hat{h}(G_{t,j}  \otimes  {G}_{{pro}}))}{nm}\\
\end{split}
\end{equation}}
Using Eq.\ref{eq:norm}, we can obtain $s(\hat{h}(G_{s,i}), \hat{h}(G_{s,j}))\leq \sqrt{k_{G}}(k_{G}-1)L_{\hat{E}}\mu$ and similarly, $s(\hat{h}(G_{t,i}), \hat{h}(G_{t,j}))\leq \sqrt{k_{G}}(k_{G}-1)L_{\hat{E}}\mu$. 
Then, we derive Eq.\ref{eq:upperbound_main} by incorporating these inequalities into Eq.\ref{eq:dd_main}. We can further derive the upper bound of the backdoor learning task $\mathcal  L^{t}_{\text{bdk}}$ in the downstream application context using the loss $\mathcal L^{s}_{\text{bdk}}$ of the backdoor learning task at the pretraining stage in Eq.\ref{eq:dd_bdk_target}:
{\begin{equation}\label{eq:dd_bdk_target}
\small
\begin{split}
&\mathcal L^{t}_{\text{bdk}} \leq \mathcal L^{s}_{\text{bdk}} + 2\sqrt{d(G_{s}\oplus\Delta_{G},(G_{t,j}\oplus\Delta_{G} )  \otimes  {G}_{{pro}})} + C_{1}\\
&d(G_{s}\oplus\Delta_{G},(G_{t,j}\oplus\Delta_{G})  \otimes  {G}_{{pro}})\\
&= \sum\limits_{i,j=1}^{n}\frac{s(\hat{h}(G_{s,i}\oplus\Delta_{G}),\hat{h}(G_{s,j}\oplus\Delta_{G}))}{n^2} \\
&+ \sum\limits_{i,j=1}^{m}\frac{s(\hat{h}((G_{t,i}\oplus\Delta_{G})\otimes{G}_{{pro}}),\hat{h}((G_{t,j}\oplus\Delta_{G})\otimes{G}_{{pro}}))}{m^2}\\
&- \sum\limits_{i=1}^n\sum\limits_{j=1}^m \frac{s(\hat{h}(G_{s,i}\oplus\Delta_{G}),\hat{h}((G_{t,j}\oplus\Delta_{G} ) \otimes {G}_{{pro}}))}{nm}\\
&\leq 2\sqrt{k_{G}}(k_{G}-1)L_{\hat{E}}\mu-\sum\limits_{i= 1}^n \sum\limits_{j = 1}^m \frac{s(\hat{h}(G_{s,i}\oplus\Delta_{G} ),\hat{h}((G_{t,j}\oplus\Delta_{G} )  \otimes  {G}_{{pro}}))}{nm}\\ 
\end{split}
\end{equation}
By combing Eq.\ref{eq:upperbound_main} and Eq.\ref{eq:dd_bdk_target} together, we can derive Eq.\ref{eq:upperbound_bdk}}.

\noindent\textbf{Proof to Proposition \ref{proposition1}.}
For the simplicity of analysis, we consider prompts as prompt graph patterns $G_{{pro}}$ attached to an input graph $G$. We can then view the prompted graph $G \otimes G_{{pro}}$ as the result of applying an isolated component transformation $g_{\text{ict}}$ to the input graph $G$.  

\begin{proposition}\label{prop:prompt}
\textbf{Proposition 5 in \cite{DBLP:journals/corr/abs-2209-15240}.} For a frozen graph encoder $E$, we assume each node of an input graph $G$ has a $\text{vdim}$-dimensional feature vector. There hence exists a node-wise feature perturbation $\delta\mathrm{feat}(V)\in{R^{\text{vdim}}}$ applied to the node features of $G$ (noted as $\mathrm{feat}(V)$), which satisfies: 
\begin{equation}
\small
\begin{split}
&E(G\otimes G_{\text{pro}}) = E(G')\\
&G' = (\mathrm{feat}(V) + \delta\mathrm{feat}(V),A) \\
\end{split}
\end{equation}
where $A$ is the graph adjacency matrix of $G$. 
\end{proposition}

Building on Proposition \ref{prop:prompt}, we can deduce the analytical form of $\delta\mathrm{feat}(V)$ by assuming the encoder $E$ employs a GIN architecture. We follow the assumption originally established in the proof to Proposition \ref{prop:prompt}, the GIN structure has a linear layer for node feature extraction and a sum readout function. Therefore, the node embedding produced by $E$ can be given as: 
\begin{equation}
\small
H=(A+(1+\epsilon))\mathrm{feat}(V)\theta
\end{equation}
where $\theta$ denotes the parameters of the linear layer in the GIN-based graph encoder. The graph embedding of $G$ is then derived as $SUM(H)$, by adding up the node embedding of each node $V\in{G}$. 
Hence, we can given the analytical expression of $\delta\mathrm{feat}(V)$ as below, following Appendix A.3 in \cite{DBLP:journals/corr/abs-2209-15240}:
\begin{equation}
\small
\delta\mathrm{feat}(V)^{i} = \frac{\sum_{j=1}^{k_{{pro}}}(A_{{pro}}+(1+\epsilon)I)\mathrm{feat}(V_{{pro}})^{i}_{j}}{\text{Deg}+N(1+\epsilon)} 
\end{equation}
{where $\delta\mathrm{feat}(V)^{i}$ is the $i$-th element of $\delta\mathrm{feat}(V)$. The injected prompt graph has $k_{{pro}}$ nodes in total. We assume the node feature vectors in $G$ and the prompt graph $G_{{pro}}$ share the same dimension. $\mathrm{feat}(V_{{pro}})^{i}_{j}$ thus denotes the $i$-th node feature element of the node $j$ of the prompt graph. $A_{{pro}}$ is the adjacency matrix of the prompt graph $G_{{pro}}$. 
$\text{Deg}$ and $N$ are the total degree of all of the nodes and the number of graph nodes of $G$. }

In summary, embeddings of a prompted graph $G_{t,i}\otimes{G_{{pro}}}$ of the downstream application context can be represented as below. Let $V(G_{t,i})$ are the nodes of $G_{t,i}$. 
$\delta\mathrm{feat}(V(G_{t,i}))$ is the additional perturbation over the node features of $G_{t,i}$. $\mathrm{feat}(V_{pro})$ denotes the node features of the prompt graph $G_{{pro}}$.
{\begin{equation} 
\small
\begin{split}
&H^{t}(G_{t,i}\oplus{G_{{pro}}}) = (A_{t,i} + (1 + \epsilon))(\mathrm{feat}(V(G_{t,i})) + \delta\mathrm{feat}(V(G_{t,i})))\theta\\
&\delta\mathrm{feat}(V(G_{t,i})) = [\frac{\sum_{j=1}^{k_{\text{pro}}}(A_{\text{pro}}+(1+\epsilon)I)\mathrm{feat}(V_{pro})^{0}_{j}}{\text{Deg}_{t,i}+N_{t,i}(1+\epsilon)}, \\
&\frac{\sum_{j=1}^{k_{{pro}}}(A_{{pro}}+(1+\epsilon)I)\mathrm{feat}(V_{pro})^{1}_{j}}{\text{Deg}_{t,i}+N_{t,i}(1+\epsilon)},...,\\
&\frac{\sum_{j=1}^{k_{{pro}}}(A_{{pro}}+(1+\epsilon)I)\mathrm{feat}(V_{pro})^{\text{vdim}}_{j}}{\text{Deg}_{t,i}+N_{t,i}(1+\epsilon)} ]\\
\end{split}
\end{equation}}

Similarly, for an input graph $G_{s,i}$ at the pretraining stage, let $V(G_{s,i})$ denote the set of nodes in $G_{s,i}$. Embeddings of $G_{s,i}$ produced by the encoder are given as:
\begin{equation}
\small
H^{s}(G_{s,i}) = (A_{s,i} + (1 + \epsilon))\mathrm{feat}(V(G_{s,i})) \theta
\end{equation}

Assuming the prompt graph $G_{{pro}}$ always has a fully connected structure, tuning the node features of the prompt graph $\mathrm{feat}(V_{pro})$ to minimize 
$\sum\limits_{i= 1}^n \sum\limits_{j = 1}^m \frac{s(\hat{h}(G_{s,i}),\hat{h}(G_{t,j}  \otimes  {G}_{{pro}}))}{nm}$ in Eq.\ref{eq:upperbound_main} is equivalent to solving the following least squared regression problem:
{\begin{equation}\label{eq:lsr}
\small
\begin{split}
&\mathrm{feat}^{*}(V_{pro}) = \underset{\mathrm{feat}(V_{pro})}{\arg\,\min}\sum_{i=1}^{n}\sum_{j=1}^m \|\text{SUM}((A_{s,i} + (1 + \epsilon))\mathrm{feat}(V(G_{s,i}))\theta)-\\
&\text{SUM}((A_{t,j} + (1 + \epsilon))(\mathrm{feat}(V(G_{t,j})) + \delta\mathrm{feat}(V(G_{t,j})))\theta)\|^{2}_{2}\\
\end{split}
\end{equation}}
where $\|\|_{2}$ is the $L$-2 norm. Therefore, for any set of the graphs $\{G_{s,i}\}$ and $\{G_{t,j}\}$ for pretraining and the downstream learning tasks, there exists an optimal $\mathrm{feat}^{*}(V_{pro})$ solving the optimization problem in Eq.\ref{eq:lsr}. 

\noindent\textbf{Proof of Proposition \ref{proposition2}.}
Given a graph $\Delta_{G}$ as the trigger graph in \textit{CrossBA}, an augmented graph $\Delta_{G}^+$ is derived by adding / removing links from $\Delta_{G}$. The backdoored graph $G\oplus{\Delta_G}$ is generated by linking the trigger graph $\Delta_G$ to an anchor node in an input graph $G$. 

\begin{proposition}\label{prop:prompt2}
\textbf{Proposition 4 in \cite{DBLP:journals/corr/abs-2209-15240}.} We perform a link transformation over a graph $G$, i.e. adding / removing links from a graph $G$ containing $k_{G}$ nodes, to derive an augmented graph $G^{+}$. For a frozen graph encoder $E$, we assume each node of an input graph $G$ has a $\text{vdim}$-dimensional feature vector. There hence exists a node-wise feature perturbation $\delta\mathrm{feat}(V)\in{R^{\text{vdim}}}$ applied to the node features of $G$ (noted as $\mathrm{feat}(V)$), which satisfies:
{\begin{equation}
\small
\begin{split}
&H(G^{+}) = (A+(1+\epsilon)I)(\mathrm{feat}(V(G)) + \delta\mathrm{feat}(V(G^{+})))\theta\\
&\delta\mathrm{feat}(V)^{k} = \frac{\sum_{i,j=1}^{k_{G}}\delta{A}_{i,j}\mathrm{feat}(V(G))_{j}^k}{\text{Deg}+(1+\epsilon)N}\\  
\end{split}
\end{equation}}
where $A$ is the graph adjacency matrix of $G$. ${\delta}{A}_{i,j}$ are the difference between the graph adjacency of $G$ and $G^{+}$, i.e. ${\delta}{A}_{i,j} = A_{i,j} - A^{+}_{i,j}$. $\mathrm{feat}(V)_{j}^k$ ($j=1,2,3...,k_{G}$ and $k=1,2,3,...,\text{vdim}$) denotes the $k$ dimension of the feature vector of node $j$. 
\end{proposition}

Using Proposition 5 in \cite{DBLP:journals/corr/abs-2209-15240}, the node embedding of the backdoored graph $G_{s,i} \oplus \Delta_{G}$ can be expressed as: 
{\begin{equation}
\small
\begin{split}
&H(G_{s,i}\oplus\Delta_{G}) = (A_{s,i} + (1 + \epsilon))(\mathrm{feat}(V(G_{s,i})) + \delta\mathrm{feat}(V(G_{s,i})))\theta\\
&\delta\mathrm{feat}(V(G_{s,i})) = [\frac{\sum_{j=1}^{k_{\text{trigger}}}(A_{\text{trigger}}+(1+\epsilon)I)\mathrm{feat}(V(\Delta_G))^{0}_{j}}{\text{Deg}_{s,i}+N_{s,i}(1+\epsilon)}, \\
&\frac{\sum_{j=1}^{k_{\text{trigger}}}(A_{\text{trigger}}+(1+\epsilon)I)\mathrm{feat}(V(\Delta_G))^{1}_{j}}{\text{Deg}_{s,i}+N_{s,i}(1+\epsilon)},...\\
&,\frac{\sum_{j=1}^{k_{\text{trigger}}}(A_{\text{trigger}}+(1+\epsilon)I)\mathrm{feat}(V(\Delta_G))^{\text{vdim}}_{j}}{\text{Deg}_{s,i}+N_{s,i}(1+\epsilon)}]\\
\end{split}
\end{equation}} \normalsize
where $k_{\text{trigger}}$, $A_{\text{trigger}}$ and $\mathrm{feat}(\Delta_G)$ are the number of the nodes, the adjacency matrix and the node feature matrix of the trigger graph $\Delta_G$. $\text{Deg}_{s,i}$ and $N_{s,i}$ are the total degree and the number of nodes in the graph $G_{s,i}$.

In summary, tuning $\mathrm{feat}(V(\Delta_G))$,  the node features of the trigger graph, to minimize $\frac{1}{nm}{\sum\limits_{i= 1}^n \sum\limits_{j = 1}^m s(\hat{h}(\Delta^{+}_{G,i}),\hat{h}(G_{s,j} \oplus \Delta_{G}))}$ can be formulated as a least squared regression problem in Eq.\ref{eq:lsr2}. For any set of $\Delta^{+}_{G,i}$ and $G_{s,j}$, there exists an optimal $\mathrm{feat}^{*}V(\Delta_G)$ solving the optimization problem in Eq.\ref{eq:lsr2}. 
{\begin{equation}\label{eq:lsr2}
\small
\begin{split}
&\mathrm{feat}^{*}(V(\Delta_G)) =\\
&\underset{\mathrm{feat}(V(\Delta_G))}{\arg\min}\,\,\sum_{i=1}^n\sum_{j=1}^m\|\text{SUM}((A+(1+\epsilon)I)(\mathrm{feat}(V(G)) + \delta\mathrm{feat}(V(G^{+}_{i})))\theta) - \\
&\text{SUM}((A_{s,i} + (1 + \epsilon))(\mathrm{feat}(V(G_{s,i})) + \delta\mathrm{feat}(V(G_{s,i})))\theta)\|^{2}_{2}\\
\end{split}
\end{equation}}

\section{Datasets and GNN Models}
\label{app:datasets}

CiteSeer \cite{yang2016revisiting} comprises 3,312 scientific publications categorized into 6 classes, connected by 4,732 citation links. Cora \cite{yang2016revisiting} consists of 2,708 scientific publications, each assigned to one of 7 categories within a citation network with 5,429 links. Amazon-Computers and Amazon-Photo \cite{shchur2018pitfalls} are subgraphs of the Amazon co-purchase graph, where nodes represent products, and edges indicate frequent co-purchases. Amazon-Computers has 13,752 nodes across 10 categories and 491,722 edges, while Amazon-Photo comprises 7,650 nodes from 8 categories and 238,162 edges. The ENZYMES dataset \cite{borgwardt2005protein} contains 600 enzymes from the BRENDA enzyme database, classified into 6 EC enzyme categories. For graph classification datasets derived from node classification datasets, we follow the methodology proposed in \cite{allinone}, involving edge and subgraph sampling from the original data. Detailed statistics are presented in Table \ref{tab:datasets}, where the last column indicates the type of downstream task for each dataset: "N" for node classification and "G" for graph classification.

GAT \cite{GAT} and GT \cite{GT} are two advanced GNN models. GAT employs neighborhood aggregation for node embedding learning and distinguishes itself by assigning varied weights to neighboring nodes, thereby modifying their influence in the aggregation process. GT integrates the processing capabilities for graph-structured data with the self-attention mechanisms of Transformer networks, effectively capturing complex relationships and feature dependencies between nodes in a graph. 

\begin{table}[t]
\caption{Statistics of datasets.}
\label{tab:datasets}
\resizebox{\columnwidth}{!}{%
\begin{tabular}{c|c|c|c|c|c}
\toprule
Datasets & Nodes & Edges  & Features & Labels & Tasks \\ 
\midrule
CiteSeer    &  3,327       & 9,104 &   3,703   &6& N   \\ 
Cora    &   2,708       & 5,429  &   1,433  &7&  N   \\ 
Amazon-Computers  &    13,752      & 491,722 &   767   &10&  N  \\ 
Amazon-Photo    &    7,650      & 491,722 &  767    &8& N   \\ 
ENZYMES   &    32.63(Avg.)       &  62.14(Avg.) &  18 &6& G  \\ 
\bottomrule
\end{tabular}
}
\end{table}

\section{Baseline Attacks}\label{app:baselineattack}
GCBA \cite{GCBA} is the most relevant backdoor attack method to our study, given the threat model definition. Unlike \textit{CrossBA}, which targets the GPL framework, GCBA aims to inject backdoor poisoning noise into a GNN encoder trained using Graph Contrastive Learning (GCL).
The attacker in GCBA can collect the graph data of the target class in the downstream applications. The attack is then formulated to maximize the similarity between the embeddings of backdoored graph data and those of clean graph data belonging to the target class in the downstream task, as well as the similarity between embeddings of clean data from both the backdoored and clean GNN encoders. However, GCBA is not directly applicable to cross-context GPL scenarios as it relies on access to downstream application data. To facilitate a fair comparison, we introduce two variants of GCBA, namely {GCBA\_R} and {GCBA\_M}. In both variants, the attacker first clusters the embeddings of clean graph data collected during pretraining using the backdoor-free GNN encoder. Then, GCBA\_R randomly selects an embedding centered around a cluster in the embedding space as the target embedding, while GCBA\_M selects the cluster center furthest away from other class centers as the target embedding. Subsequently, both variants follow the same workflow as GCBA to execute the attack. 

Compared to \textit{CrossBA}, GCBA differs in two significant aspects in its algorithmic design. Firstly, while \textit{CrossBA} optimizes the target embedding during training of the backdoored GNN encoder (as shown in Eq. \ref{eq:optimise}), GCBA uses a fixed target embedding derived from the downstream dataset's target class. 
Secondly, \textit{CrossBA} includes a node feature affinity constraint in its attack objective, aiding in circumventing countermeasures that inspect node feature outliers, such as PruneG. However, GCBA permits arbitrary node features in the injected trigger graph. Consequently, \textit{CrossBA} presents a more stealthy attack compared to GCBA, particularly against countermeasures deployed by downstream users.

\section{Implementation Details}\label{app:implementation}
We implement \textit{CrossBA} using PyTorch and execute it on an NVIDIA 3090 GPU.
For both GAT and GT, we adopt a two-layer graph neural network structure with a hidden dimension set to 100. Following the methodology of ProG \cite{allinone}, we utilize Singular Value Decomposition (SVD) to reduce the initial feature dimension of the data to 100. The number of prompt nodes used in ProG and ProG-Meta is set to 15. The self-supervised learning method employed by ProG and ProG-Meta is GraphCL \cite{graphcl}. Consistent with the original paper's settings, we employ the Adam optimizer for ProG and ProG-Meta. On most evaluated datasets, the learning rate for GT is set to 0.001, and for GAT, it is set to 0.0001. Similarly, following the settings of GraphPrompt \cite{graphprompt}, we use the AdamW optimizer with a unified learning rate of 0.01 for all test instances. The self-supervised learning method used by GraphPrompt is the link prediction method \cite{graphprompt}. Additionally, for downstream tasks, we employ 200-shot learning for ProG and ProG-Meta, and 100-shot learning for GraphPrompt.

In this study, we investigate five distinct cross-context scenarios. To address cross-distribution, cross-dataset, and cross-domain scenarios, we adopt the methodology of ProG \cite{allinone}, which leverages clustering methods to divide the entire graph into 200 distinct subgraphs for creating the pretraining dataset. For downstream tasks, we follow ProG's approach to construct induced graph datasets tailored for both node and graph classification tasks. 
In cross-class scenarios, we initially divide the label space into pretraining classes and downstream classes. Subsequently, we partition the entire graph into two disjoint subgraphs according to these classes, which are then used to form induced graph datasets for pretraining and downstream tasks, respectively.
For the cross-task scenario, we utilize either the real graph classification dataset ENZYMES or the generated induced graph dataset for graph classification tasks as the pretraining dataset. The induced graph dataset for node classification tasks is employed as our downstream dataset.

Regarding the backdoor attack setting, for all the attack methods, the trigger graph is designed to consist of only three nodes, significantly fewer than the number of nodes in the input graph. During pretraining, the attacker randomly selects a node in the input graph as the anchor node. We set $\alpha=0.5$ for both the baseline attack methods and \textit{CrossBA}. Additionally, for \textit{CrossBA}, we set $\lambda$ and $\beta$ to 0.05 across the majority of datasets. The Adam optimizer is used for optimizing the trigger graph and the backdoored GNN encoder. Specifically, for \textit{CrossBA}, we set $\gamma_t$ to 0.01 and $\gamma_g$ to 0.0001.  For GCBA\_R and GCBA\_M, following the setting in \cite{GCBA}, we set $\gamma_t$ to 0.0015 and $\gamma_g$ to 0.001.

\section{Attack Results in Cross-dataset and Cross-task Scenarios}\label{app:attackresult}
\begin{table*}[htbp]
\centering
\footnotesize
\caption{ACC, ASR, and AD in cross-dataset scenarios. Cora is used as the pretraining dataset for CiteSeer and CiteSeer-Graph, and Computers is used for Photo and Photo-Graph. }
\label{tab:cross_dataset}
\begin{tabular}{@{}ccc *{8}{|c}@{}}
\toprule
& & & \multicolumn{4}{c|}{\textbf{Node Classification}} & \multicolumn{4}{c}{\textbf{Graph Classification}} \\ 
\cmidrule(lr){4-7} \cmidrule(lr){8-10}
\multirow{2}{*}{\textbf{GPL}} & \multirow{2}{*}{\textbf{Model}} & \multirow{2}{*}{\textbf{Attack}} & \multicolumn{2}{c|}{\textbf{CiteSeer}}  & \multicolumn{2}{c|}{\textbf{Photo}} & \multicolumn{2}{c|}{\textbf{CiteSeer-Graph}} & \multicolumn{2}{c}{\textbf{Photo-Graph}} \\ 
\cmidrule(lr){4-5} \cmidrule(lr){6-7} \cmidrule(lr){8-9} \cmidrule(lr){10-11} 
 & & & ACC(AD) & ASR & ACC(AD) & ASR & ACC(AD) & ASR& ACC(AD) & ASR \\ 
\midrule
\multirow{6}{*}{ProG} & \multirow{3}{*}{GAT}& GCBA\_R   & 0.24(+0.57) & 0.48 & 0.44(+0.34)& 0.59 & 0.17(+0.61) & 0.00 & 0.59(+0.30) & 0.28 \\
                          &                           &GCBA\_M    & 0.19(+0.62) &0.00 & 0.46(+0.32) & 0.28 & 0.17(+0.61) & 0.06 & 0.44(+0.45) & 0.68 \\
                          && CrossBA & \textbf{0.83(-0.02)}& \textbf{1.00}& \textbf{0.78(-0.00)} & \textbf{1.00} & \textbf{0.79(-0.01)} & \textbf{1.00} & \textbf{0.89(-0.00)} & \textbf{0.97}  \\
\cmidrule(lr){2-11}
                          & \multirow{3}{*}{GT}       & GCBA\_R   & 0.48(+0.34) & 0.80 & 0.55(+0.26) & 0.95 & 0.44(+0.35) & 0.42 & 0.67(+0.27) & 0.99 \\
                          &                           & GCBA\_M   & 0.48(+0.34) & 0.90 & 0.52(+0.29)& 0.16 &0.44(+0.35) & 0.75 &0.58(+0.36) & 0.62 \\
                          & & CrossBA     & \textbf{0.82(-0.00)} & \textbf{1.00} & \textbf{0.81(-0.00)} & \textbf{1.00}& \textbf{0.79(-0.00)} & \textbf{1.00} & \textbf{0.93(+0.01)} &  \textbf{1.00}  \\
\midrule
\multirow{6}{*}{\begin{tabular}[c]{@{}l@{}}Graph\\Prompt\end{tabular}}  & \multirow{3}{*}{GAT}  & GCBA\_R   & \textbf{0.72(-0.00)} & 0.09 &  0.60(+0.12) &0.12 &   0.67(+0.06) &0.10 & 0.71(+0.10) & 0.10 \\
                          &                           & GCBA\_M   & 0.59(+0.13) & 0.12 & 0.64(+0.08)& 0.03 & 0.64(+0.09) & 0.03& 0.65(+0.16)& 0.16 \\
                          && CrossBA & 0.67(+0.05) &\textbf{0.97} & \textbf{0.70(+0.02)} &\textbf{1.00} & \textbf{0.67(+0.06)} & \textbf{0.93} & \textbf{0.82(-0.01)} &\textbf{ 1.00}  \\
\cmidrule(lr){2-11}
                          & \multirow{3}{*}{GT}      & GCBA\_R   & 0.66(+0.13) & 0.06& 0.59(+0.13) & 0.13 &  0.59(+0.18) &0.04 &0.68(+0.16) & 0.09 \\
                          &                           & GCBA\_M   & 0.55(+0.24) & 0.01 & 0.58(+0.14) & 0.12 &  0.66(+0.11)& 0.09&0.66(+0.18) &  0.32 \\
                          & & CrossBA     & \textbf{0.80(-0.01)} &\textbf{1.00} & \textbf{0.71(+0.01)} &  \textbf{0.99} & \textbf{0.77(-0.00)} & \textbf{1.00} & \textbf{0.85(+0.01)} &  \textbf{0.93}  \\
\midrule[0.75pt] 
\multirow{6}{*}{\begin{tabular}[c]{@{}l@{}}ProG\\Meta\end{tabular}}   & \multirow{3}{*}{GAT} & GCBA\_R   & 0.50(+0.17) & 0.00 & 0.83(+0.14) & 1.00 & 0.50(+0.29) & 0.00 & 1.00(-0.00) &  0.96\\
                          &                           & GCBA\_M   & 0.50(+0.17) & 0.00 &  0.50(+0.47) & 0.00 & 0.50(+0.29) & 0.00& 0.75(+0.25) & 1.00 \\
                          && CrossBA & \textbf{0.92(-0.25)} & \textbf{1.00} & \textbf{0.97(-0.00)} &\textbf{1.00} & \textbf{0.89(-0.10)} & \textbf{1.00 }& \textbf{1.00(-0.00)} &\textbf{1.00} \\
\cmidrule(lr){2-11}
                          & \multirow{3}{*}{GT}       & GCBA\_R   & 0.94(-0.00) & 1.00 & 0.89(+0.10) & 0.91 & 0.50(+0.44) & 0.00 & 1.00(-0.00) & 0.01 \\
                          &                           & GCBA\_M   & 0.71(+0.23) & 0.29 & 0.55(+0.44) & 0.00 & 0.68(+0.26) & 1.00 & 0.95(+0.05) & 0.96 \\
                          && CrossBA     & \textbf{0.94(-0.00)} & \textbf{1.00}& \textbf{0.98(+0.01)} & \textbf{1.00} & \textbf{0.92(+0.02)} &\textbf{1.00 }& \textbf{1.00(-0.00) }&\textbf{1.00 } \\
\bottomrule
\end{tabular}
\end{table*}

\begin{table}[htbp]
\centering
\footnotesize
\caption{ACC, ASR, and AD in cross-task scenarios. ENZYMES is the pretraining dataset.}
\label{tab:cross_level}
\resizebox{\columnwidth}{!}{%
\begin{tabular}{@{}ccc|c|c|c|c@{}}
\toprule
\multirow{2}{*}{\textbf{GPL}} & \multirow{2}{*}{\textbf{Model}} & \multirow{2}{*}{\textbf{Attack}} & \multicolumn{2}{c|}{\textbf{CiteSeer}} & \multicolumn{2}{c}{\textbf{Cora}} \\ 
\cmidrule(lr){4-5} \cmidrule(lr){6-7}
 & & & ACC(AD)  & ASR  & ACC(AD)  & ASR \\ 
\midrule
\multirow{6}{*}{ProG} & \multirow{3}{*}{GAT} & GCBA\_R   & 0.50(+0.26) & 0.23 & 0.24(+0.42) &  0.00 \\
                       &                      & GCBA\_M   & 0.61(+0.15) & 0.10 & 0.24(+0.42) & 0.00 \\
                       && CrossBA      & \textbf{0.75(+0.01)} &\textbf{ 1.00} & \textbf{0.72(-0.06)} & \textbf{1.00} \\
\cmidrule(lr){2-7}
                       & \multirow{3}{*}{GT} & GCBA\_R   &  0.55(+0.15) & 0.85 & 0.44(+0.20) & 0.80 \\
                       &                      & GCBA\_M   &  0.57(+0.13) & 0.77 & 0.44(+0.20) & 0.93 \\
                       && CrossBA     &\textbf{ 0.82(-0.12)} &\textbf{ 0.87} &\textbf{ 0.64(-0.00)} & \textbf{1.00} \\
\midrule
\multirow{6}{*}{\begin{tabular}[c]{@{}l@{}}Graph\\Prompt\end{tabular}} & \multirow{3}{*}{GAT} &GCBA\_R   &  0.70(+0.04) &  0.03 & 0.26(+0.38) & 0.41 \\
                             &                      & GCBA\_M   & 0.71(+0.03) &  0.00 & 0.34(+0.30) &0.13 \\
                             && CrossBA & \textbf{0.75(-0.01)} & \textbf{1.00} & \textbf{0.65(-0.01)} &  \textbf{0.97}\\
\cmidrule(lr){2-7}
                             & \multirow{3}{*}{GT}  & GCBA\_R   &  0.69(+0.06)& 0.05 &  \textbf{0.56(-0.03)} & 0.24 \\
                             &                      & GCBA\_M   & \textbf{0.77(-0.02)} & 0.04 & 0.50(+0.03)& 0.26 \\
                             && CrossBA     & 0.75(-0.00) & \textbf{ 1.00 }& 0.53(-0.00) & \textbf{1.00}\\
\midrule[0.75pt]
\multirow{6}{*}{\begin{tabular}[c]{@{}l@{}}ProG\\Meta\end{tabular}} & \multirow{3}{*}{GAT} & GCBA\_R   & 0.50(+0.14) &0.00 & 0.50(+0.38) & 0.00\\
                           &                      & GCBA\_M   & 0.76(-0.12) & 1.00 & 0.50(+0.38)& 0.00 \\
                           && CrossBA &\textbf{ 0.90(-0.26)} &\textbf{1.00} & \textbf{0.88(-0.00)} & \textbf{0.87} \\
\cmidrule(lr){2-7}
                           & \multirow{3}{*}{GT} & GCBA\_R   & 0.91(+0.03) & 0.61 & 0.50(+0.41) &0.00 \\
                           &                      & GCBA\_M   & 0.89(+0.05) & 0.44 &  0.51(+0.40) & 1.00  \\
                           && CrossBA     & \textbf{0.93(+0.01)} & \textbf{1.00}& \textbf{0.91(-0.00)} & \textbf{1.00} \\
\bottomrule
\end{tabular}
}
\end{table}



Table \ref{tab:cross_dataset} highlights the attack performance of various backdoor attacks in cross-dataset scenarios, where Cora is used as the pretraining dataset for CiteSeer and CiteSeer-Graph, while Computers serves as the pretraining dataset for Photo and Photo-Graph. The results underscore the superior attack capabilities of \textit{CrossBA} across diverse downstream applications. \textit{CrossBA} consistently achieves ASR values surpassing 0.83 in all tested cases, with a maximum decrease of only 0.06 in ACC compared to the clean model. In contrast, baseline methods encounter challenges in transferring backdoors to downstream models without significantly compromising the main task's performance.


In cross-task scenarios, the GNN encoder model is pretrained on ENZYMES, consisting of actual molecular graphs. The downstream node classification tasks involve CiteSeer and Cora. From the results in Table \ref{tab:cross_level}, we find that \textit{CrossBA} effectively injects backdoors into downstream models pretrained on the graph classification task while maintaining high ACC values in node classification tasks. Conversely, the baseline methods often face challenges in transferring backdoors to downstream node classification tasks. For example, when targeting GraphPrompt, the baseline methods struggle to achieve an ASR higher than 0.41. In contrast, \textit{CrossBA} consistently achieves ASR values surpassing 0.97.

\section{Attack against PruneG}\label{app:defense}

\begin{figure}[t] 
\centering  
\includegraphics[height=3cm,width=8cm]{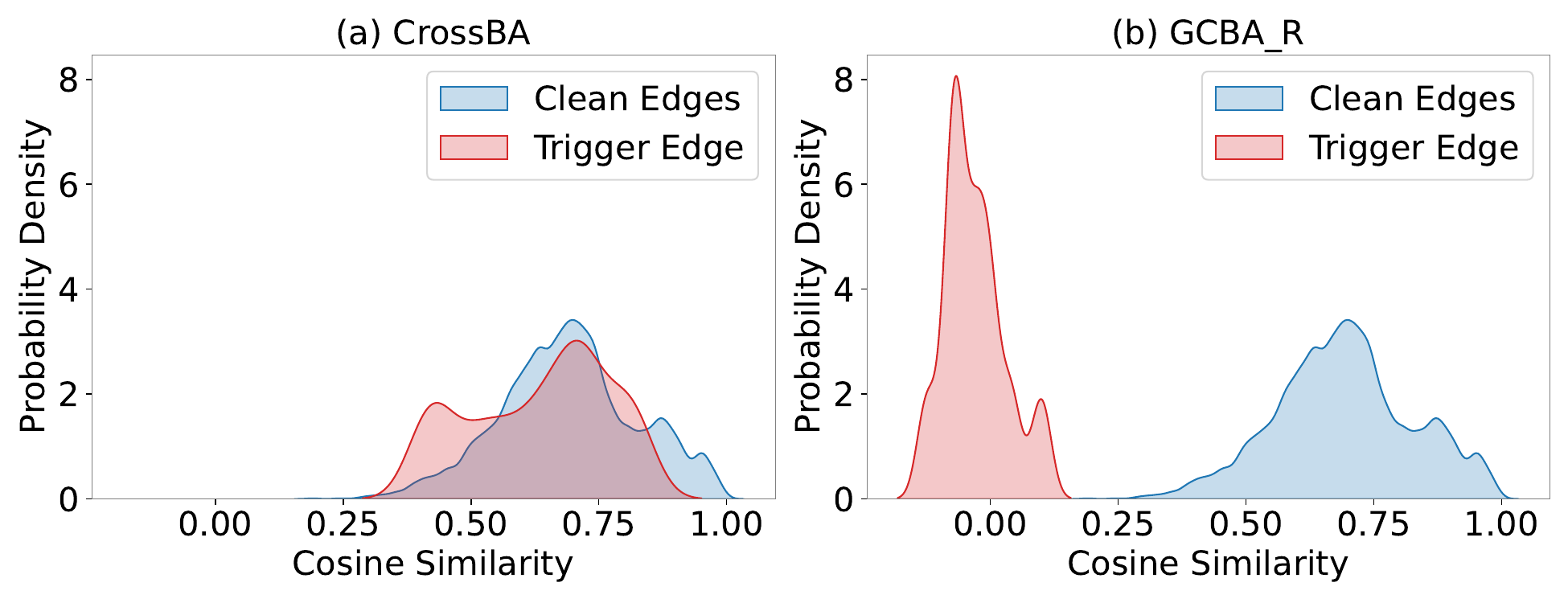} \caption{Kernel density estimation of node feature similarity on CiteSeer. } 
\label{fig:node_sim}
\end{figure}

We explore potential countermeasures against the \textit{CrossBA} attack. Given the lack of specific defenses for cross-context GPL scenarios, we adapt defenses from other scenarios to mitigate the \textit{CrossBA} attack. Real-world graphs, such as social networks, typically exhibit homophily, where nodes with similar features are connected by edges. However, trigger graph injection can disrupt this homophily. Therefore, a category of defense methods exists that improves GNN robustness by pruning edges connecting nodes with low feature similarity \cite{wu2019adversarial}. We extend the \textit{PruneG} defense to align with our threat model, where the downstream user acts as the defender. Given a graph, the defender calculates the similarity of node features connected by edges and prunes those edges with similarity below the threshold, removing the component with fewer nodes that the edge connects.

Figure \ref{fig:defense} shows the \textit{PruneG} defense's effectiveness against backdoor attacks in both node and graph classification tasks across 5 cross-context scenarios and 2 GPL methods. Cora is employed for pretraining in cross-dataset scenarios, Photo serves as the pretraining dataset in cross-domain scenarios, and the induced graph dataset for the graph classification task is used as the pretraining dataset in cross-task scenarios. 
All the baseline methods exhibit poor attack performance facing the \textit{PruneG} defense. In contrast, \textit{CrossBA} successfully evades detection, achieving high ASR values. For example, in cross-domain scenarios, \textit{CrossBA} attains ASR values above 0.90 on all the tested cases, while the baselines' ASR values remain below 0.50. In \textit{CrossBA}, the optimization of trigger node features is constrained to closely resemble those of the clean anchor nodes. 
Figure \ref{fig:node_sim} illustrates the kernel density of node feature similarity values, highlighting that trigger nodes created by the baseline attacks exhibit minimal similarity to anchor nodes' features, making them more easily detected. Conversely, \textit{CrossBA} achieves indistinguishable similarity for trigger edges compared to clean edges, confirming its stealthiness.

\section{Ablation study}\label{app:ablation}
\begin{figure*}[t] 
\centering  
\includegraphics[height=3cm,width=17cm]{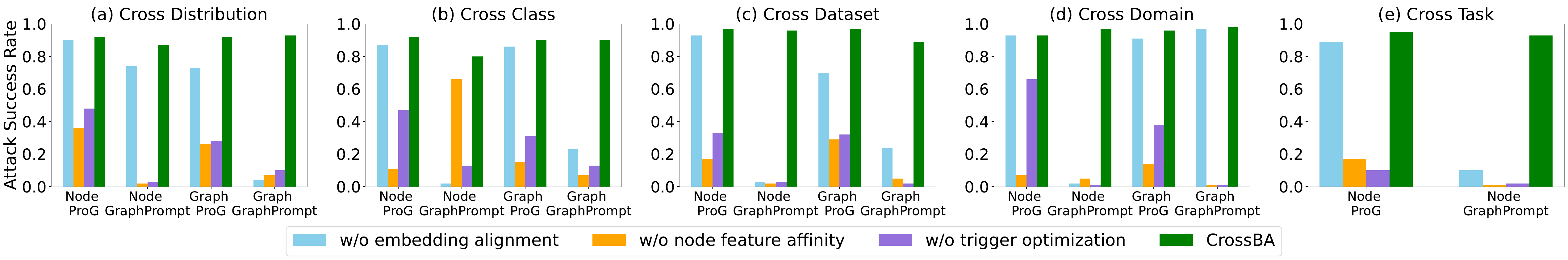} \caption{Ablation study of CrossBA against 2 GPL methods with Prune on CiteSeer across 5 cross-context scenarios. "Node" represents the node classification task, and "Graph" denotes the graph classification task. } 
\label{fig:ablation}
\end{figure*}

\begin{figure*}[t] 
\centering  
\includegraphics[height=3.5cm,width=17.5cm]{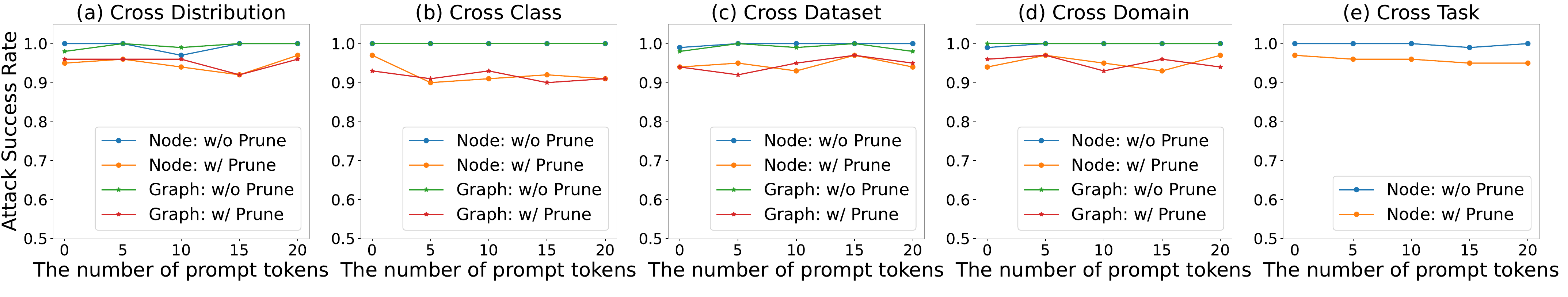} \caption{Impact of the number of prompt tokens on attack performance of CrossBA against ProG based on CiteSeer in 5 cross-context scenarios. "Node" represents the node classification task, and "Graph" denotes the graph classification task. } 
\label{fig:prompt node}
\end{figure*}

\begin{figure*}[t] 
\centering  
\includegraphics[height=4.25cm,width=17.5cm]{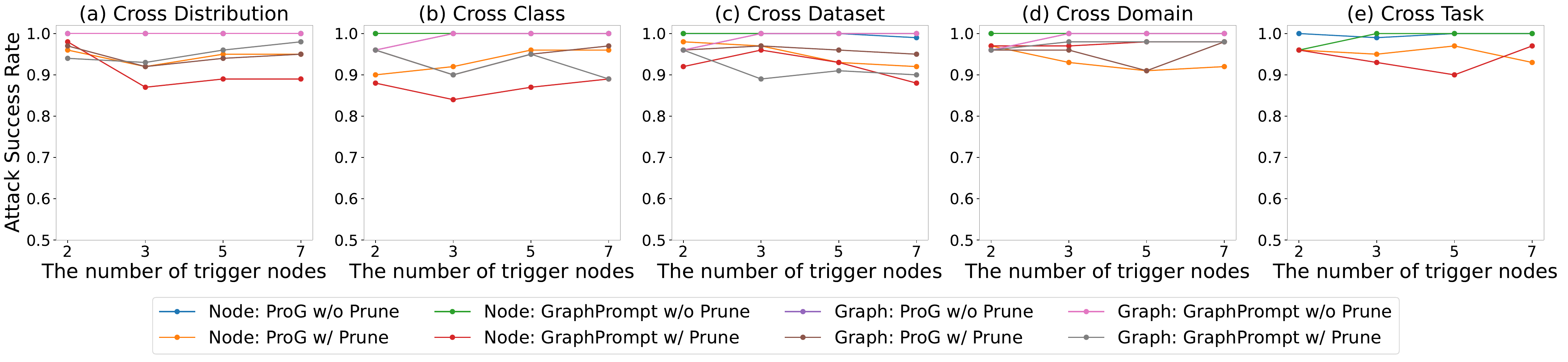} \caption{Impact of the number of trigger nodes on attack performance of CrossBA against ProG and GraphPrompt based on CiteSeer in 5 cross-context scenarios. "Node" represents the node classification task, and "Graph" denotes the graph classification task.} 
\label{fig:trigger node}
\end{figure*}

In our ablation studies on \textit{CrossBA} to assess the significance of its components, we consider three variants: (1) \textit{CrossBA} without trigger optimization, employing a fixed trigger graph and target embedding; (2) \textit{CrossBA} without embedding alignment ($\alpha=0$); (3) \textit{CrossBA} without node feature affinity ($\beta=0$).
Figure \ref{fig:ablation} illustrates the attack performance of \textit{CrossBA} and its variants against two GPL methods with the Prune defense, encompassing five cross-context scenarios. Cora is used for pretraining in cross-dataset scenarios, Photo serves as the pretraining dataset in cross-domain settings, and the induced graph dataset for the graph classification task is employed in cross-task scenarios.

Comparing \textit{CrossBA} to its three variants, we observe that \textit{CrossBA} consistently maintains stable attack performance across various cross-context GPL scenarios, even in the presence of defense mechanisms. Each component of \textit{CrossBA} is crucial for achieving high ASR values. For instance, in cross-task scenarios involving Graphprompt, the removal of any component significantly lowers ASR values to below 0.20. Conversely, \textit{CrossBA} attains an ASR of 0.90, underscoring the significance of each component to the attack's success.

\section{Impact of Prompt Tokens.} \label{app:prompttoken}
Figure \ref{fig:prompt node} illustrates the impact of the number of prompt tokens on the attack performance of \textit{CrossBA} against ProG in both node and graph classification tasks on CiteSeer across five cross-context scenarios. In cross-dataset scenarios, Cora is employed for pretraining, while Photo serves as the pretraining dataset for cross-domain settings. In cross-task scenarios, the generated induced graph dataset for the graph classification task is utilized for pretraining.

The results reveal that the effectiveness of \textit{CrossBA} remains robust, exhibiting little impact from variations in the number of prompt tokens. Across all five cross-context scenarios, \textit{CrossBA} consistently maintains an ASR above 0.90, even with an increased number of prompt tokens. Importantly, this stability is observed regardless of the presence of defense mechanisms, emphasizing the resilience of \textit{CrossBA} to variations in prompt token numbers.

\section{Impact of Trigger Nodes.}\label{app:triggernodes}
Figure \ref{fig:trigger node} illustrates the impact of varying the number of trigger nodes on the attack performance of \textit{CrossBA} against ProG and GraphPrompt in both node and graph classification tasks on CiteSeer across five cross-context scenarios. In cross-dataset scenarios, the pretraining is conducted using Cora, while Photo serves as the pretraining dataset for cross-domain settings. In cross-task scenarios, the generated induced graph dataset for the graph classification task is utilized for pretraining.
The results highlight that \textit{CrossBA} consistently maintains stable attack performance, achieving an ASR above 0.80 across all the test cases, regardless of the specific number of trigger nodes employed.

\end{document}